\pdfoutput=1
\documentclass[11pt]{article}

\usepackage[final]{acl}

\usepackage{times}
\usepackage{latexsym}

\usepackage[T1]{fontenc}
\usepackage[utf8]{inputenc}

\usepackage{microtype}

\usepackage{inconsolata}

\usepackage{graphicx}
\usepackage{algorithm}
\usepackage{algorithmic}
\usepackage{booktabs}
\usepackage{xspace}
\usepackage{enumitem}

\newcommand{\algo}{HATA\xspace}
\newcommand{\topk}{top-$k$}

\newcommand{\textbt}[1]{\textbf{\textit{#1}}}

\definecolor{darkgreen}{rgb}{0.078,0.667,0.016}

\usepackage{amsmath}
\usepackage{amssymb}
\usepackage{mathtools}
\usepackage{amsthm}

\usepackage{bm}
\usepackage{multirow}
\usepackage{multicol}
\usepackage{subcaption}
\usepackage{colortbl}
\usepackage{stfloats}
\usepackage{comment}
\usepackage{tablefootnote}

\title{HATA: Trainable and Hardware-Efficient Hash-Aware Top-$k$ Attention for Scalable Large Model Inference}

\author{
 \textbf{Ping Gong\textsuperscript{\dag,\ddag,\S}},
 \textbf{Jiawei Yi\textsuperscript{\dag, \ddag}},
 \textbf{Shengnan Wang\textsuperscript{\S}},
 \textbf{Juncheng Zhang\textsuperscript{\dag}},
 \textbf{Zewen Jin\textsuperscript{\dag}},
\\
 \textbf{Ouxiang Zhou\textsuperscript{\dag}},
 \textbf{Ruibo Liu\textsuperscript{\dag}},
 \textbf{Guanbin Xu\textsuperscript{\dag}},
 \textbf{Youhui Bai\textsuperscript{\S}},
 \textbf{Bowen Ye\textsuperscript{\P}},
 \textbf{Kun Yuan\textsuperscript{\P}},
\\
 \textbf{Tong Yang\textsuperscript{\P}},
 \textbf{Gong Zhang\textsuperscript{\S}},
 \textbf{Renhai Chen\textsuperscript{\S}},
 \textbf{Feng Wu\textsuperscript{\dag,\ddag}},
 \textbf{Cheng Li\textsuperscript{\dag,\ddag}}
\\
 \textsuperscript{\dag}University of Science and Technology of China,
\\ 
 \textsuperscript{\ddag}Institute of Artificial Intelligence, Hefei Comprehensive National Science Center,
\\
 \textsuperscript{\S}Huawei Technologies,
 \textsuperscript{\P}Peking University
\\
   \href{mailto:gpzlx1@mail.ustc.edu.cn}{gpzlx1@mail.ustc.edu.cn}, \href{mailto:chengli7@ustc.edu.cn}{chengli7@ustc.edu.cn}
}

\begin{document}
\maketitle

\renewcommand{\thefootnote}{\fnsymbol{footnote}}
\footnotetext[1]{Cheng Li is the corresponding author.}
\footnotetext[1]{Work done during Ping's internship at Huawei.}
\footnotetext[1]{Ping and Jiawei equally contributed to this work.}
\renewcommand{\thefootnote}{\arabic{footnote}}

\begin{abstract}

Large Language Models (LLMs) have emerged as a pivotal research area, yet the attention module remains a critical bottleneck in LLM inference, even with techniques like KVCache to mitigate redundant computations. While various top-$k$ attention mechanisms have been proposed to accelerate LLM inference by exploiting the inherent sparsity of attention, they often struggled to strike a balance between efficiency and accuracy. In this paper, we introduce \textbf{HATA (Hash-Aware Top-$k$ Attention)}, a novel approach that systematically integrates low-overhead learning-to-hash techniques into the Top-$k$ attention process. Different from the existing top-k attention methods which are devoted to seeking an absolute estimation of qk score, typically with a great cost, HATA maps queries and keys into binary hash codes, and acquires the relative qk score order with a quite low cost, which is sufficient for realizing top-k attention. 
Extensive experiments demonstrate that HATA achieves up to \textbf{7.2$\times$ speedup} compared to vanilla full attention while maintaining model accuracy. In addition,  HATA outperforms the state-of-the-art top-$k$ attention methods in both accuracy and efficiency across multiple mainstream LLM models and diverse tasks. HATA is open source at \href{https://github.com/gpzlx1/HATA}{https://github.com/gpzlx1/HATA}.

\end{abstract}
\section{Introduction}
\label{intro}

Recently, KVCache has become a paradigm for the inference of large language model (LLM) \cite{vllm, SGLang}, due to its benefit of mitigating redundant computation in the decoding stage. In this situation, massive KV states loading becomes the bottleneck, especially for long 
sequences and large batch sizes~\cite{sparq, quest}.

\begin{figure}[!t]
  \centering
    \centering
    \includegraphics[width=.40\textwidth]{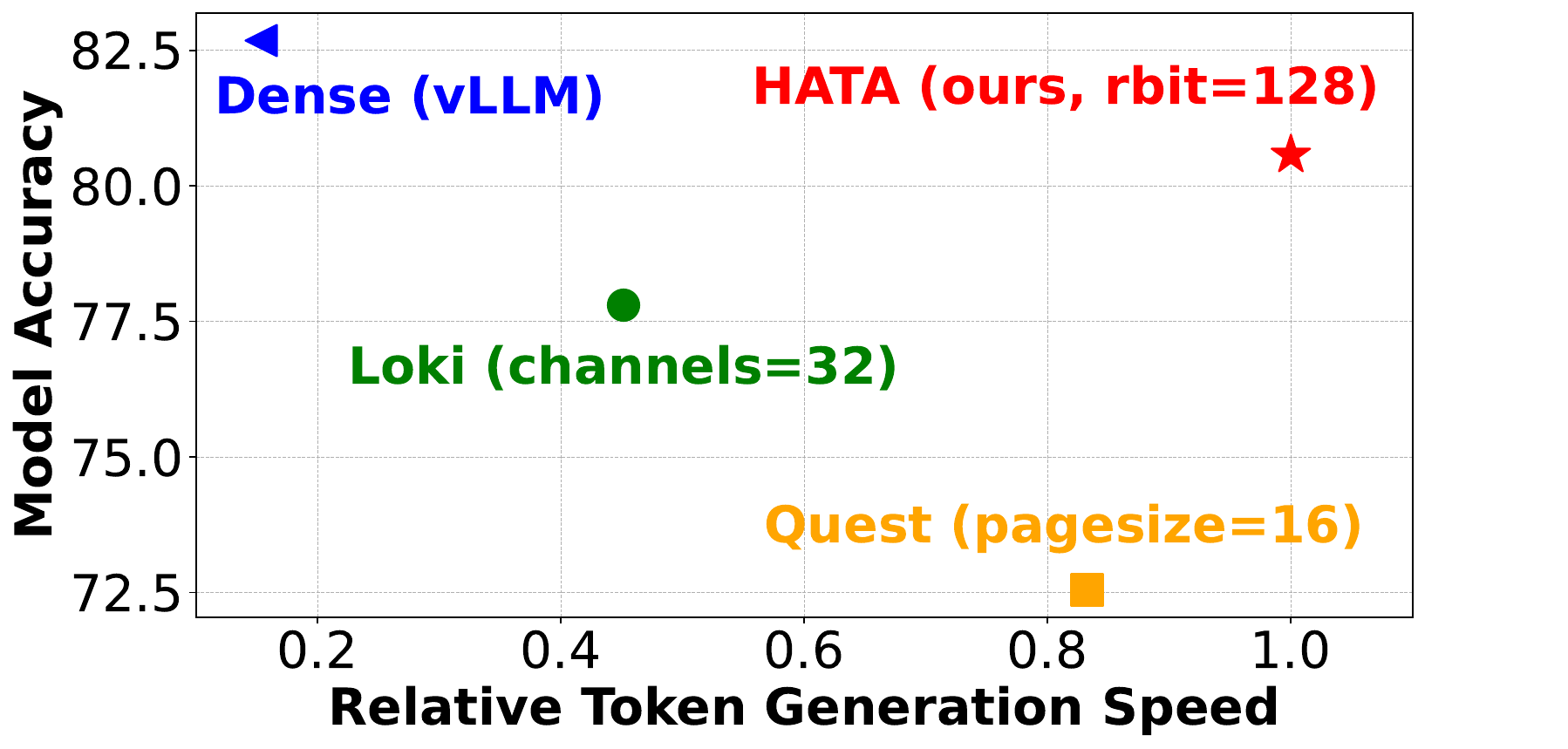}
    \caption{Comparison of accuracy and token generation speed. For detailed analysis, refer to Sec~\ref{sec:evaluations}.}
    \label{fig:accuracy and performance}
\end{figure}
\begin{comment}
\end{comment}

\textbf{Top-$k$ Attention}~\cite{topk-attention} has emerged as a promising approach to accelerate LLM inference by leveraging the inherent sparsity in attention. By selectively retaining only the top-\(k\) most relevant tokens in the KVCache, top-\(k\) attention significantly reduces the KVCache loading overhead. However, existing top-\(k\) attention algorithms face notable challenges in achieving an optimal trade-off between efficiency and accuracy. Low-rank methods, such as Loki~\cite{loki} and InfiniGen~\cite{infinigen}, reduce overhead by computing dot-products over a subset of projected dimensions, but they introduce significant computational costs due to the extensive requirements for channel extraction. On the other hand, block-wise methods like Quest~\cite{quest} and InfLLM~\cite{infllm} improve efficiency by grouping contiguous key-value pairs into blocks, but they often compromise accuracy as critical keys may be excluded based on their coarse-grained estimation of query-key (qk)  scores.

In this paper, we introduce \textbf{Hash-Aware Top-\(k\) Attention (HATA)}, a novel approach that systematically integrates low-overhead learning-to-hash techniques into the top-\(k\) attention process. Unlike existing methods that focus on precise numerical estimation of qk scores, HATA maps queries and keys into binary hash codes, acquiring the relative qk score order with minimal computational cost. This approach eliminates costly high-fidelity score approximations, enabling significant speedup while preserving the quality of top-\(k\) selection. As illustrated in Figure~\ref{fig:accuracy and performance}, \algo shows superiority in balancing the accuracy and efficiency, compared to state-of-the-art methods.

HATA leverages the success of \textbf{learning-to-hash}~\cite{wang2012semi,weiss2008spectral}, which has been widely used in similarity-based retrieval tasks such as image search and machine learning. By training hash functions based on the query-key pairs of LLM attention, HATA is able to encode any query and key vector into a binary code, which further enable HATA to achieve  low-overhead but  precise token selection, making it a hardware-efficient solution for accelerating LLM inference.

Extensive experiments demonstrate that HATA achieves up to 7.2\(\times\) speedup compared to vanilla full attention while maintaining model accuracy. Furthermore, HATA outperforms state-of-the-art top-\(k\) attention methods in both accuracy and efficiency across multiple mainstream LLM models and diverse tasks.

In summary, our contributions are as follows:

\begin{itemize}[noitemsep, topsep=2pt]
    \item We frame key retrieval in top-$k$ attention as a lightweight ordinal comparison task, eliminating the need for costly high-fidelity score approximation.

    \item We introduce \algo,  which systematically integrates learning-to-hash techniques into top-$k$ attention mechanisms to solve this ordinal comparison task.

    \item We provide hardware-aware optimizations for \algo and validate its effectiveness on multiple models and datasets.
\end{itemize}

\section{Background and Motivation}
\label{sec:backgroun and motivation}

\subsection{LLM Inference}

The LLM model consists of multiple transformer layers, each processing continuous token embeddings to iteratively generate the next token embedding. At the core of each transformer layer is the attention module, which computes as follows:

\begin{equation}
\begin{aligned}
  Q, K, V &= \text{Proj}\left( X \right), \\
  AttnOut &= \text{Softmax}\left(\frac{QK^T}{\sqrt{d}}\right)V.
\end{aligned}
\label{eq:topk_attention }
\end{equation}

LLM inference is autoregressive. When generating text, the model produces one token at a time, and each new token depends on the ones already generated. This process continues until some stopping condition is met, like reaching an end-of-sequence token or a maximum length. %%CL generating each token based on all preceding context until the target output sequence is complete. 
However, the autoregressive nature leads to significant computational redundancy, making attention the primary bottleneck in LLM inference~\cite{flash-attention, flash-attention-2, you2024linearattention}. 

\subsection{KVCache} 

To accelerate the attention module, the KVCache approach has been proposed to cache and reuse intermediate results to eliminate computational overhead.

In more detail, it decouples  the inference process into \textbf{prefill} and \textbf{decode} stages. During the \textit{prefill stage}, the input prompt is processed in parallel, computing and caching the \textit{K} and \textit{V} vectors for all tokens across the transformer-attention layers, which initializes the KVCache. 
In the subsequent \textit{decode stage}, tokens are generated sequentially: at each step, the model computes only the Q/K/V vector of the current token, retrieves cached K/V vectors, and computes attention scores to predict the next token, while appending the new token’s K/V vectors to the cache.

Despite the KVCache's computational efficiency, the attention mechanism remains a critical bottleneck for modern LLMs in complex scenarios involving long-context sequences or large batch sizes. Recent studies~\cite{sparq, quest} reveal that even with KVCache, the attention module dominates inference latency—for instance, consuming over 70\% of total runtime when processing 32K-token sequences with Llama2-7B. This inefficiency is contributed not only by the computation complexity but also by memory bandwidth constraints. At each decoding step, the model must load the entire cached Key and Value vectors, incurring massive data movement costs that scale with context length and batch size. Consequently, with KVCache, optimizing attention’s memory access patterns has emerged as a pivotal challenge for enabling scalable LLM deployment.

\subsection{Top-$k$ Attention}

The top-$k$ attention mechanism~\cite{topk-attention} reduces memory bandwidth overhead under the KVCache framework by exploiting the sparsity of attention distributions. As formalized in Equation~(\ref{eq:topk}), it computes attention scores only for the top-$k$ keys with the highest query-key (qk) scores, bypassing computation for low-scoring tokens. While this sparsity preserves model accuracy and reduces FLOPs, it does not fully eliminate the memory bottleneck: as shown in~\cite{sparq}, the mechanism still requires loading all keys from the KVCache to evaluate qk scores, incurring at least half of the original memory traffic.

To improve the efficiency of top-$k$ attention, recent work has focused on approximating qk scores with low-cost estimators. Methods like SparQ~\cite{sparq}, Loki~\cite{loki}, and InfiniGen~\cite{infinigen} reduce computational overhead by computing dot-products over a subset of projected dimensions rather than the full embedding space. While these approximations retain theoretical error bounds, they face a dimensionality-accuracy trade-off: preserving estimation fidelity requires retaining a critical mass of dimensions, leading to limited performance gains.

\begin{equation}\label{eq:topk}
\begin{aligned}
    qkScore &= \text{Softmax}(qK^T) \\
    Index &= \text{TopK}(qkScore, k)  \\
    AttnOut &= \text{Attn}(q, K[Index], V[Index])
\end{aligned}
\end{equation}

On the other side, block-based approximations, such as Quest~\cite{quest} and InfLLM~\cite{infllm}, partition keys into contiguous blocks and estimate upper bounds for aggregate qk scores per block. Tokens within blocks exceeding a score threshold are retained for attention computation. While this reduces the search space, two issues arise. Critical tokens are often dispersed across blocks, and selecting entire blocks forces loading irrelevant intra-block keys, wasting memory bandwidth. Moreover, the coarse-grained estimation may not well distinguish important and irrelevant tokens, hindering the final accuracy.

\subsection{Motivation}

Prior top-$k$ attention methods operate under the strong assumption that precise numerical estimation of qk scores is essential to replicate the effectiveness of full attention. Thus, they incur significant computational or memory overhead to minimize approximation errors in absolute qk scores. 

However, in this paper, we challenge this assumption by demonstrating that only relative qk score ordering---not absolute numerical magnitude---is required to identify the most relevant keys. By reformulating the problem as a lightweight \textit{ordinal comparison} task (e.g., determining whether $s_{qk_{i}} > s_{qk_{j}}$) rather than a \textit{numerical regression} task, we eliminate the need for costly high-fidelity score approximations. This relaxation enables remarkable reduction in computation and memory access while preserving top-$k$ selection quality, as precise score magnitudes are irrelevant to the ranking outcome.

Learning-to-hash~\cite{wang2012semi} offers a principled framework to achieve our goal, as it maps high-dimensional continuous vectors (e.g. queries and keys) into compact binary hash codes while preserving their relative similarity relationships, i.e., similar vectors are assigned adjacent binary hash codes with small Hamming distances.

Nevertheless, integrating learning-to-hash into top-$k$ attention introduces critical challenges:  
\begin{itemize}[noitemsep, topsep=2pt]

	\item \textbf{Modeling.} Learning-to-hash was widely used for retrieval tasks, such as image retrieval and information search. To apply learning-to-hash to top-$k$ attention computing, designing an effective hashing model for learning hash codes of query and keys is of great importance.

	\item \textbf{Implementation.} A high-performance implementation is also indispensable to achieve a practical improvement of LLM inference. 
\end{itemize}	
\section{\algo's Design}
\label{sec: Learning-to-hash and Design}
To address the aforementioned three challenges, we propose Hash-Aware Top-$k$ Attention (HATA), a trainable and hardware-efficient approach based on learning-to-hash.

In Sec~\ref{sec:problem formluation}, we formally define the query-key-based learning-to-hash problem and design a training loss function to learn hash codes while preserving similarity. We also incorporate bits balance and uncorrelation constraints~\cite{wang2012semi, weiss2008spectral} to enhance hash bit quality. In Sec~\ref{sec:algorithm}, we introduce HATA's workflow, leveraging the learned hash function to significantly accelerate LLM inference.

\subsection{Learning-to-Hash for Top-$k$ Attention}
\label{sec:problem formluation}

Building on learning-to-hash, we design a hash function to map query/key vectors to binary codes while preserving their relative similarity. The learning process is detailed below.

\subsubsection{Hash Modeling}
Inspired by the learning-to-hash model defined in~\cite{wang2012semi}, given a query $q$ and multiple keys  $K:=\{k_i\}_{i=1}^n$, 
we learn the hash codes of $q$ and $K$ by solving the following problem:
\begin{eqnarray}
    \min &&\sum_{i} \text{sim}(q, k_i) ||h(q) - h(k_i)||^2 \label{eq:obj} \\
    \text{s.t.} && h(q),h(k_i) \in \{-1,1\}^r  \label{eq:encode} \\
    && \sum_{i=1}^n  h(k_i) = 0 \label{eq:balance}\\ 
    && \frac{1}{n}\sum_{i=1}^n h(k_i)h(k_i)^T  = I_r \label{eq:indep} 
\end{eqnarray}
where $h(\cdot)$ is the hash function to be learned and $\text{sim}(q, k_i)$ defines the similarity of original query $q$ and key $k_i$. Note that the objective function Equation (\ref{eq:obj}) tends to assign adjacent binary codes for qk pairs exhibiting high similarity,  which matches the goal of similarity-preserving hashing. The constraint (\ref{eq:encode}) ensures that the query and keys are encoded into $r$ binary codes. The constraints (\ref{eq:balance}) and (\ref{eq:indep}) are called bits balance and uncorrelation constraints, respectively. %%CL, which are widely used in learning-to-hash to make the binary codes efficient.

The hash function $h(\cdot)$ is commonly defined as $h(x) = \text{sign}(xW_H)$, where $W_H$ is the trainable hash weights. 

Due to the non-differentiability of the sign function, we relax $h(x)$ as:
\begin{equation}
   h(x) = 2 \cdot \text{Sigmoid}(\sigma \cdot xW_H) - 1,
\end{equation}
where $\sigma\in(0,1)$ is a hyper-parameter to prevent gradient vanishing.

For tractability, the balance constraint  (\ref{eq:balance}) is further relaxed by minimizing $||\sum_{i}h(k_i)||^2$, and according to~\cite{wang2012semi} the uncorrelation constraint (\ref{eq:indep}) can be relaxed by minimizing $||W_{H}^{T}W_H - I_r||$. Then 
the query-key hashing problem is reformulated as:
\begin{align}
    \min &&&\epsilon\sum_{i} s_i ||h(q) - h(k_i)||^2 + \notag\\ 
         &&& \eta ||\sum_{i}h(k_i)||^2 + \lambda||W_H^TW_H - I_r|| \label{eq:optim_obj}\\
    \text{s.t.} &&&h(x) = 2 \cdot \text{Sigmoid}(\sigma \cdot xW_H) - 1 \nonumber
\end{align}
where $s_i$ is $\text{sim}(q, k_i)$ for short, and \(\epsilon, \lambda, \eta\) control the impact of each objective. Detailed training settings are provided in the Appendix~\ref{appndx:hashtrain:settings}.

For convenience, we first formulate the learing-to-hash problem based on a single query and its corresponding keys, as shown in Equation~(\ref{eq:optim_obj}). This objective function consists of three components. The first term, $\min \sum_{i} s_i ||h(q) - h(k_i)||^2$, serves as the main objective, enforcing similarity preservation by ensuring that similar items maintain close hash codes in the binary hash space. The terms $\min ||\sum_{i}h(k_i)||^2$ and $\min||W_H^TW_H - I_r||$ further ensure the efficiency of the learned hash codes. Next, we extend Equation~(\ref{eq:optim_obj}) to a more realistic case that includes multiple queries, as below:
\begin{align}
    \min &&&  \epsilon\sum_{j}\sum_{i} s_{j,i} ||h(q_j) - h(k_{j,i})||^2 + \notag\\ 
         &&& \eta \sum_{j}||\sum_{i}h(k_{j,i})||^2 + \lambda||W_{H}^{T}W_H - I_r|| \label{eq:qkfinal}\\
    \text{s.t.} &&&h(x) = 2 \cdot \text{Sigmoid}(\sigma \cdot xW_H) - 1 \nonumber
\end{align}
where $s_{j,i} = \text{sim}(q_j,k_i).$ Problem (\ref{eq:qkfinal}) is the final hashing model for learning effective hash function $h(\cdot)$, which plays a key role in designing efficient top-$k$ attention algorithm later.

Note that the attention module typically involves multiple independent heads which usually have different characteristics, so we also train a separate hash weight $W_H$ for each attention head.

\subsubsection{Training Data Construction}

The training samples are constructed based on real datasets.
Specifically, given a sequence, during the prefill stage, we collect $Q := [q_1, q_2, \dots, q_n]$ and $K := [k_1, \dots, k_n]$ of each attention head. For each head, we sample a $q_j$ from $Q$ and compute the qkScore between $q_j$ and $[k_1, \dots, k_j]$. Based on the qkScore, the top 10\% of $(q_j k_i)$ pairs are designated as positive samples with linearly decayed labels $s_{j,i} \in [1, 20]$, while the remaining 90\% receive fixed negative labels $s_{j,i} = -1$. The label $s_{j,i}$ measures the similarity between $q_j$ and $k_i$. The training data are organized as triplets $(q_j, k_i, s_{j,i})$ for storage. Since the sequence can be very long, it is easy to generate thousands or even millions of qk pairs for training. To enhance data diversity, we generate training data from dozens of sequences. The details of this process are presented in Appendix~\ref{appndx:hashtrain:data}.

\begin{algorithm}[t]
   \caption{\algo Prefill Stage}
   \label{alg:hash-attn-prefill}
\begin{algorithmic}[1]
   \STATE {\bfseries Input:} \textbt{Q} $\in \mathbb{R}^{s\times d}$, \textbt{K} $\in \mathbb{R}^{s\times d}$, \textbt{V} $\in \mathbb{R}^{s\times d}$, key cache $\bm{K^{cache}}$ $\in \mathbb{R}^{0\times d}$, value cache $\bm{V^{cache}}$$ \in \mathbb{R}^{0\times d}$, key code cache $\bm{K^{cache}_H}$ $\in \mathbb{R}^{0\times rbit/32}$
   \STATE \textcolor{darkgreen}{$\triangleright$ Call HashEncode to encode key}
   \STATE $\bm{K_H}$ $ \leftarrow $ HashEncode(\textbt{K})
   \STATE \textcolor{darkgreen}{$\triangleright$ Fill hashcode cache}
   \STATE $\bm{K^{cache}_H}$ $\leftarrow$ $\bm{K_H}$ %, $\bm{K^{cache}_N}$ $\leftarrow$ $\bm{K_N}$
   \STATE \textcolor{darkgreen}{$\triangleright$ Fill KVCache}
   \STATE $\bm{K^{cache}}$ $\leftarrow$ $\textbt{K}$, $\bm{V^{cache}}$ $\leftarrow$ $\textbt{V}$
   \STATE \textcolor{darkgreen}{$\triangleright$ Calculate attention output}
   \STATE \textbt{O} $\leftarrow$ Attention(\textbt{Q}, \textbt{K}, \textbt{V})
\end{algorithmic}
\end{algorithm}

\subsection{\algo Top-$k$ Attention Algorithm}
\label{sec:algorithm}
\algo integrates learning-to-hash to top-$k$ attention via two algorithmic innovations: (1) HATA Prefill: caching hash codes of $K$; (2) HATA Decoding: efficient top-k key-value detection through hash space.

\begin{figure*}[t]
\begin{center}
\centerline{\includegraphics[width=\columnwidth*2]{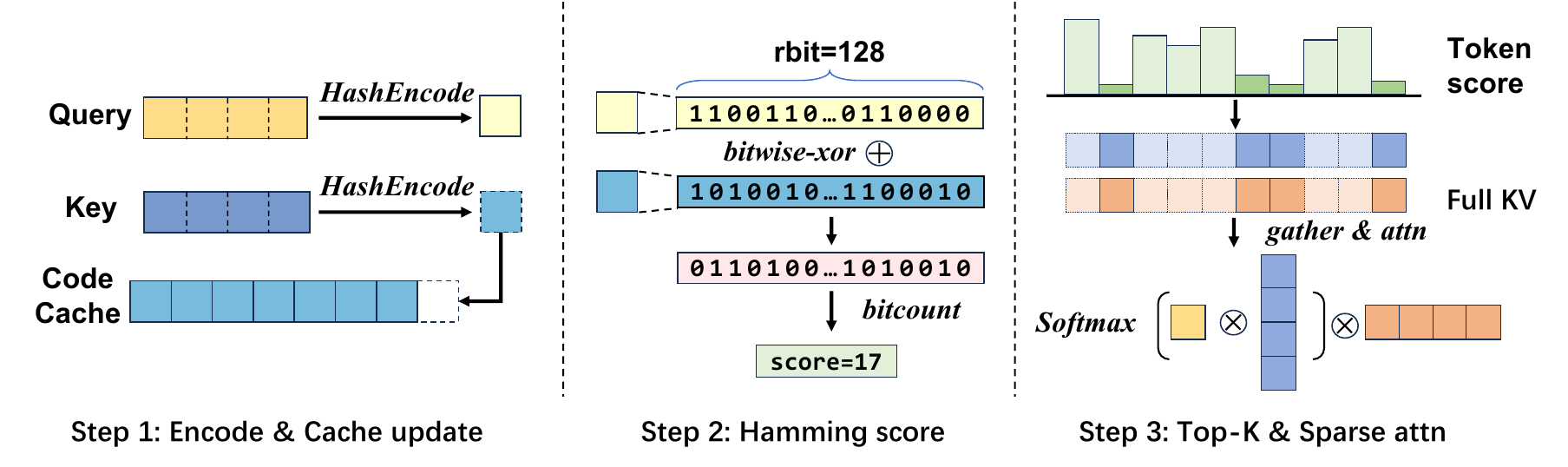}}
\caption{Workflow of \algo in the decode stage.}
\label{fig:hash-decode}
\end{center}
\end{figure*}

\begin{algorithm}[t]
   \caption{HashEncode}
   \label{alg:hash-encode}
\begin{algorithmic}[1]
   \STATE {\bfseries Input:} vector \textbt{V} $\in \mathbb{R}^{s\times d}$
   \STATE {\bfseries Parameter:} hash weight $\bm{W_H}$$ \in \mathbb{R}^{d\times rbit}$
   \STATE {\bfseries Output:} hash code $\bm{V_H}$ $\in \mathbb{N}^{s\times rbit/32}$
   \STATE \textcolor{darkgreen}{$\triangleright$ Project input vector into hash code}
   \STATE $\bm{V_H}$ $ \leftarrow $ Sign(MatMul(\textbt{V}, $\bm{W_H}$))
   \STATE \textcolor{darkgreen}{$\triangleright$ Pack hash code bits into integer format}
   \STATE $\bm{V_H}$ $\leftarrow$ BitPack($\bm{V_H}$)
   
\end{algorithmic}
\end{algorithm}

\begin{algorithm}[t]
   \caption{\algo Decode Stage}
   \label{alg:hash-attn-decode}
\begin{algorithmic}[1]
   \STATE {\bfseries Input:} \textbt{Q} $\in \mathbb{R}^{1\times d}$, \textbt{K} $\in \mathbb{R}^{1\times d}$, \textbt{V} $\in \mathbb{R}^{1\times d}$, key cache $\bm{K^{cache}}$ $\in \mathbb{R}^{s\times d}$, value cache $\bm{V^{cache}}$ $ \in \mathbb{R}^{s\times d}$, key code cache $\bm{K^{cache}}_H$ $\in \mathbb{R}^{s\times rbit/32}$, top-$k$ number $N$
   \STATE \textcolor{darkgreen}{$\triangleright$ Update KVCache}
   \STATE $\bm{K^{cache}}$ $\leftarrow$ $[\bm{K^{cache}}; \textbt{K}]$
   \STATE $\bm{V^{cache}}$ $\leftarrow$ $[\bm{V^{cache}}; \textbt{V}]$
   \STATE \textcolor{darkgreen}{$\triangleright$ Call HashEncode to encode query and key}
   \STATE $\bm{Q_H}$ $ \leftarrow $ HashEncode(\textbt{Q})
   \STATE $\bm{K_H}$ $ \leftarrow $ HashEncode(\textbt{K})
   \STATE \textcolor{darkgreen}{$\triangleright$ Update code cache with $K_H$}
   \STATE $\bm{K^{cache}_H}$ $\leftarrow$ $[\bm{K^{cache}_H}; \bm{K_H}]$
   % \STATE $\bm{K^{cache}_N}$ $\leftarrow$ $[\bm{K^{cache}_N}; \bm{K_N}]$
   \STATE \textcolor{darkgreen}{$\triangleright$ Calculate distance in Hamming space}
   \STATE \textbt{S} $\leftarrow$ bitcount(bitwise\_xor($\bm{Q_H}$,$\bm{K^{cache}_H}$)) 
   \STATE \textcolor{darkgreen}{$\triangleright$ Select top-$k$ key-value pairs}
   \STATE \textbt{Idx} $\leftarrow$ TopK(\textbt{S}, $N$)
   \STATE $\bm{K^{sparse}}$ $\leftarrow$ Gather($\bm{K^{cache}}$, \textbt{Idx})
   \STATE $\bm{V^{sparse}}$ $\leftarrow$ Gather($\bm{V^{cache}}$, \textbt{Idx})
   \STATE \textcolor{darkgreen}{$\triangleright$ Calculate sparse attention output}
   \STATE \textbt{O} $\leftarrow$ Attention(\textbt{Q}, $\bm{K^{sparse}}$, $\bm{V^{sparse}}$)
\end{algorithmic}
\end{algorithm}

\noindent\textbf{\algo prefill stage.} 

As shown in Algorithm~\ref{alg:hash-attn-prefill}, \algo modifies the original prefill workflow by additionlly computing and caching the hash codes of the keys (lines 2--5), which is critical for accelerating subsequent LLM decoding stages. The hash codes are generated by \text{HashEncode}, as shown in Algorithm~\ref{alg:hash-encode}, which leverages Matmul, Sign, and BitPack operators to produce $rbit$ binary code. The hash weight $W_H$ in the \text{HashEncode} is obtained through hash training as described in Sec \ref{sec:problem formluation}. 
Note that the time complexity of $\text{HashEncode}$ is $O(s \times d \times rbit)$, where $s$ is the sequence length and $d$ is the vector dimension, while $\text{Attention}$'s complexity is $O(s^2d + s^2)$. Given $rbit \ll s$, the extra prefill overhead from \algo is negligible, accounting for less than 1\% of total computation in real tasks.

\noindent\textbf{\algo decode stage.}  
As illustrated in Algorithm~\ref{alg:hash-attn-decode} and Figure~\ref{fig:hash-decode}, \algo enhances the decode workflow with the following three steps. First, in the \textit{Encode \& Cache update} step (lines 3--9), \algo first applies $\text{HashEncode}$ to the newly generated query $Q$ and key $K$, producing query code ($Q_H$) and key code ($K_H$), and then updates the key code cache $\bm{K^{cache}_H}$. Second, it computes the qk hash scores \(S\) measured by the Hamming distances between $Q_H$ and all cached key codes in $K_H^{\text{cache}}$ (including the current $K_H$) using hardware-efficient operations: $\texttt{bitwise\_xor}$ and $\texttt{bitcount}$ (lines 10--11). In situations where multiple queries target the same KVCache, such as GQA, we additionally aggregate the scores \(S\) for shared KVCache. Third, based on the hash scores, \algo selects and gathers the most relevant keys and values (lines 13--15), which are then fed into sparse attention (line 17).

\section{Hardware-Efficient Optimizations}
\label{sec:kernel-impl}

\begin{figure}[!t]
\begin{center}
\centerline{\includegraphics[width=\columnwidth]{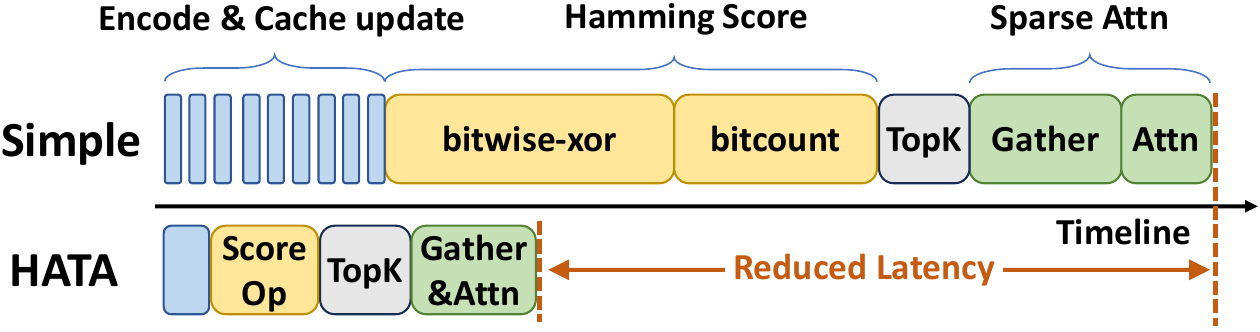}}
\caption{HATA's optimizations, compared to the conventional implementation (denoted as `Simple').}
\label{fig: kernel-optimization}
\end{center}
\end{figure}

\algo is implemented in PyTorch~\cite{pytorch} and FlashInfer~\cite{flashinfer}, comprising 1,470 lines of C++/CUDA code (for custom GPU kernels) and 940 lines of Python code (for high-level orchestration). To bridge the gap between theoretical efficiency and practical performance, we introduce three hardware-efficient optimizations, as illustrated in Figure~\ref{fig: kernel-optimization}, targeting compute and memory bottlenecks in attention with long contexts and large batches. Notably, while \algo employs extensive low-level optimizations, it remains pluggable and integrates seamlessly with existing inference frameworks. To leverage \algo, users need only replace standard attention with \algo's attention.

\noindent\textbf{Kernel fusion for hash encoding.} The \textbf{Encode \& Cache update} phase involves a chain of operations such as linear projection, sign function, BitPack, and cache updates. Although each operation takes only a few microseconds on the GPU, the CPU requires tens of microseconds to dispatch them, starving GPU compute units. By fusing these into a single CUDA kernel, we significantly reduce CPU-GPU synchronization, consequently cutting end-to-end inference latency.

\noindent\textbf{High-performance hamming score operator.} The Hamming score is computed by matching bits between query and key codes. However, PyTorch lacks high-performance operator support for this computation. To address this, we design an efficient GPU operator with the following hardware-optimized steps: First, both the query and key are loaded as multiple integers, and XOR is applied to produce intermediate integers, where `1' indicates a mismatch and `0' a match. 
% \zw{use `' for quotation marks.}
Next, the \texttt{popc}/\texttt{popcll} instructions count the number of `1's in each integer. Finally, a high-performance reduction operator aggregates these counts to compute the final score. To further boost GPU efficiency, we optimize memory bandwidth by employing coalesced memory access when transferring data from HBM to SRAM.

\noindent\textbf{Fuse gather with FlashAttention.} For \textbf{Sparse Attn}, the separate gather operations for selected keys and values result in redundant data transfers between HBM and SRAM, diminishing the benefits of hashing. To address this, we integrate the gather operation with the widely-used FlashAttention kernel~\cite{flash-attention, flash-attention-2}, streamlining data flow and reducing memory access overhead.

\section{Empirical Evaluation}
\label{sec:evaluations}

{\setlength{\tabcolsep}{2.6pt}
\begin{table*}[h]
\centering
\small
\begin{tabular}{@{}l|cccccccc|cccccccc@{}}
\toprule
\multirow{2}{*}{\textbf{Task}} & \multicolumn{8}{c|}{\textbf{Llama-2-7B-32K-Instruct}} & \multicolumn{8}{c}{\textbf{Llama-3.1-8B-Instruct}} \\
 & \textbf{Dense} & \textbf{Loki} & \textbf{Quest} & \textbf{MP} & \textbf{SL} & \textbf{H2O} & \textbf{S-KV} & \textbf{\algo} & \textbf{Dense} & \textbf{Loki} & \textbf{Quest} & \textbf{MP} & \textbf{SL} & \textbf{H2O} & \textbf{S-KV} & \textbf{\algo} \\ \midrule
\textbf{LCC} & 67.53 & 58.68 & \cellcolor[HTML]{DBFFD2}65.14 & 66.43 & 46.91 & 27.42 & 52.74 & \cellcolor[HTML]{B2ED75}\textbf{68.42} & 67.24 & 61.29 & 58.81 & 53.39 & 64.90 & 64.99 & \cellcolor[HTML]{DBFFD2}66.49 & \cellcolor[HTML]{B2ED75}\textbf{67.25} \\
\textbf{PRetr} & 11.89 & 11.97 & \cellcolor[HTML]{B2ED75}\textbf{15.53} & 10.01 & 4.84 & 2.41 & 9.44 & \cellcolor[HTML]{DBFFD2}10.61 & 99.67 & \cellcolor[HTML]{B2ED75}\textbf{99.67} & \cellcolor[HTML]{B2ED75}\textbf{99.67} & 98.83 & 94.33 & 94.00 & \cellcolor[HTML]{B2ED75}\textbf{99.67} & \cellcolor[HTML]{B2ED75}\textbf{99.67} \\
\textbf{HQA} & 15.30 & \cellcolor[HTML]{DBFFD2}14.91 & 13.64 & 14.69 & 8.22 & 4.19 & 11.78 & \cellcolor[HTML]{B2ED75}\textbf{15.65} & 60.21 & 59.48 & \cellcolor[HTML]{DBFFD2}60.03 & 56.28 & 48.52 & 57.89 & 59.93 & \cellcolor[HTML]{B2ED75}\textbf{60.19} \\
\textbf{TQA} & 85.03 & 85.30 & 85.18 & \cellcolor[HTML]{B2ED75}\textbf{86.17} & 60.67 & 20.07 & 66.9 & \cellcolor[HTML]{DBFFD2}85.83 & 91.64 & 91.45 & 90.79 & 77.90 & 79.24 & \cellcolor[HTML]{B2ED75}\textbf{92.06} & 91.98 & \cellcolor[HTML]{DBFFD2}91.94 \\
\textbf{Repo} & 55.03 & 44.41 & 52.57 & \cellcolor[HTML]{B2ED75}\textbf{55.81} & 35.43 & 16.33 & 45.13 & \cellcolor[HTML]{DBFFD2}54.92 & 52.36 & 48.47 & 46.72 & 42.35 & 45.58 & 46.70 & \cellcolor[HTML]{DBFFD2}48.6 & \cellcolor[HTML]{B2ED75}\textbf{51.72} \\
\textbf{Sam} & 39.32 & 38.95 & \cellcolor[HTML]{DBFFD2}39.24 & 38.94 & 19.50 & 6.74 & 37.56 & \cellcolor[HTML]{B2ED75}\textbf{39.61} & 42.55 & \cellcolor[HTML]{DBFFD2}41.99 & 39.75 & 34.28 & 40.10 & 41.23 & 40.58 & \cellcolor[HTML]{B2ED75}\textbf{42.35} \\
\textbf{Trec} & 69.00 & \cellcolor[HTML]{DBFFD2}69.00 & 67.57 & \cellcolor[HTML]{DBFFD2}69.00 & 28.00 & 24.33 & 37.00 & \cellcolor[HTML]{B2ED75}\textbf{69.34} & 71.66 & \cellcolor[HTML]{B2ED75}\textbf{72.33} & 71.33 & 63.67 & 51.67 & 65.00 & 60.33 & \cellcolor[HTML]{DBFFD2}71.66 \\
\textbf{MQA} & 22.44 & \cellcolor[HTML]{DBFFD2}22.11 & 19.33 & 21.70 & 15.51 & 3.06 & 17.44 & \cellcolor[HTML]{B2ED75}\textbf{22.39} & 54.82 & \cellcolor[HTML]{DBFFD2}54.47 & 51.50 & 49.10 & 34.13 & 45.12 & 52.82 & \cellcolor[HTML]{B2ED75}\textbf{55.17} \\
\textbf{2Wiki} & 13.13 & 13.09 & 12.51 & \cellcolor[HTML]{DBFFD2}13.29 & 7.54 & 2.19 & 12.65 & \cellcolor[HTML]{B2ED75}\textbf{13.44} & 44.08 & \cellcolor[HTML]{B2ED75}\textbf{44.33} & \cellcolor[HTML]{DBFFD2}43.90 & 37.84 & 37.81 & 40.91 & 43.63 & 43.82 \\
\textbf{Gov} & 32.01 & 30.51 & 24.83 & \cellcolor[HTML]{DBFFD2}31.29 & 19.39 & 9.61 & 13.54 & \cellcolor[HTML]{B2ED75}\textbf{31.90} & 35.03 & \cellcolor[HTML]{DBFFD2}34.74 & 33.64 & 32.58 & 22.73 & 29.4 & 26.29 & \cellcolor[HTML]{B2ED75}\textbf{35.02} \\
\textbf{PCnt} & 1.17 & 0.52 & \cellcolor[HTML]{B2ED75}\textbf{1.20} & \cellcolor[HTML]{DBFFD2}1.08 & 0.00 & 0.00 & 0.08 & 0.34 & 13.19 & 12.74 & \cellcolor[HTML]{B2ED75}\textbf{13.16} & 9.96 & \cellcolor[HTML]{DBFFD2}13.15 & 12.82 & 13.04 & 12.44 \\
\textbf{MltN} & 24.51 & \cellcolor[HTML]{DBFFD2}23.82 & 16.61 & 23.74 & 18.09 & 7.83 & 12.98 & \cellcolor[HTML]{B2ED75}\textbf{25.06} & 26.19 & \cellcolor[HTML]{DBFFD2}25.85 & 25.69 & 24.57 & 21.51 & 24.64 & 23.2 & \cellcolor[HTML]{B2ED75}\textbf{26.07} \\
\textbf{Qaspr} & 11.76 & \cellcolor[HTML]{B2ED75}\textbf{12.82} & 10.93 & 11.06 & 5.43 & 0.23 & 7.24 & \cellcolor[HTML]{DBFFD2}12.31 & 44.68 & \cellcolor[HTML]{B2ED75}\textbf{45.15} & 43.52 & 38.20 & 24.91 & 33.75 & 36.42 & \cellcolor[HTML]{DBFFD2}43.95 \\ \midrule
\textbf{AVG.} & 34.47 & 32.78 & 32.64 & \cellcolor[HTML]{DBFFD2}34.09 & 20.73 & 9.57 & 24.96 & \cellcolor[HTML]{B2ED75}\textbf{34.60} & 54.10 & \cellcolor[HTML]{DBFFD2}53.23 & 52.19 & 47.61 & 44.51 & 49.89 & 51.00 & \cellcolor[HTML]{B2ED75}\textbf{53.94} \\ \bottomrule
\end{tabular}

\caption{Accuracy results on LongBench-e~\cite{bai2023longbench} with 512 token budget. For MagicPIG (MP), the token budget is approximately 2-3\% of the sequence length. \textbf{SL} denotes StreamingLLM, while \textbf{S-KV} refers to SnapKV.} 
\label{tab:longbench}
\end{table*}
}

{\setlength{\tabcolsep}{2.6pt}
\begin{table*}[h]
\centering
\small
\begin{tabular}{@{}l|cccccccc|cccccccc@{}}
\toprule
\multirow{2}{*}{\textbf{Task}} & \multicolumn{8}{c|}{\textbf{Llama-2-7B-32K-Instruct}} & \multicolumn{8}{c}{\textbf{Llama-3.1-8B-Instruct}} \\
 & \textbf{Dense} & \textbf{Loki} & \textbf{Quest} & \textbf{MP} & \textbf{SL} & \textbf{H2O} & \textbf{S-KV} & \textbf{\algo} & \textbf{Dense} & \textbf{Loki} & \textbf{Quest} & \textbf{MP} & \textbf{SL} & \textbf{H2O} & \textbf{S-KV} & \textbf{\algo} \\ \midrule
\textbf{NS1} & 93.75 & 25.00 & \cellcolor[HTML]{B2ED75}\textbf{100.0} & \cellcolor[HTML]{DBFFD2}97.92 & 1.04 & 0.00 & 25.00 & \cellcolor[HTML]{B2ED75}\textbf{100.0} & 100.0 & \cellcolor[HTML]{DBFFD2}98.96 & \cellcolor[HTML]{B2ED75}\textbf{100.0} & 94.79 & 1.04 & 36.46 & \cellcolor[HTML]{DBFFD2}98.96 & \cellcolor[HTML]{DBFFD2}98.96 \\
\textbf{NS2} & 100.0 & 2.08 & \cellcolor[HTML]{DBFFD2}95.83 & \cellcolor[HTML]{DBFFD2}93.75 & 5.21 & 0.00 & 1.04 & \cellcolor[HTML]{B2ED75}\textbf{98.96} & 98.96 & \cellcolor[HTML]{DBFFD2}97.92 & 93.75 & 69.79 & 4.17 & 2.08 & 88.54 & \cellcolor[HTML]{B2ED75}\textbf{98.96} \\
\textbf{NS3} & 91.67 & 0.00 & 52.08 & 54.17 & 1.04 & 0.00 & 0.00 & \cellcolor[HTML]{B2ED75}\textbf{83.33} & 100.0 & \cellcolor[HTML]{DBFFD2}96.88 & 47.92 & 51.04 & 3.12 & 3.12 & 8.33 & \cellcolor[HTML]{B2ED75}\textbf{100.0} \\
\textbf{NMK1} & 93.75 & 0.00 & \cellcolor[HTML]{DBFFD2}87.50 & 83.33 & 6.25 & 3.12 & 2.08 & \cellcolor[HTML]{B2ED75}\textbf{93.75} & 97.92 & \cellcolor[HTML]{DBFFD2}96.88 & \cellcolor[HTML]{B2ED75}\textbf{97.35} & 65.62 & 5.21 & 4.17 & 89.58 & \cellcolor[HTML]{DBFFD2}96.88 \\
\textbf{NMK2} & 81.25 & 0.00 & 54.17 & \cellcolor[HTML]{DBFFD2}71.88 & 1.04 & 0.00 & 0.00 & \cellcolor[HTML]{B2ED75}\textbf{78.12} & 77.08 & \cellcolor[HTML]{DBFFD2}59.38 & 53.12 & 20.83 & 3.12 & 2.08 & 3.12 & \cellcolor[HTML]{B2ED75}\textbf{69.79} \\
\textbf{NMV} & 66.67 & 0.00 & 52.34 & \cellcolor[HTML]{DBFFD2}59.38 & 6.25 & 0.78 & 0.26 & \cellcolor[HTML]{B2ED75}\textbf{65.62} & 94.27 & \cellcolor[HTML]{B2ED75}\textbf{91.67} & 78.91 & 44.79 & 4.69 & 1.82 & 13.28 & \cellcolor[HTML]{DBFFD2}89.06 \\
\textbf{NMQ} & 52.08 & 0.00 & \cellcolor[HTML]{B2ED75}\textbf{56.51} & 43.49 & 4.17 & 0.00 & 0.00 & \cellcolor[HTML]{DBFFD2}54.17 & 96.09 & \cellcolor[HTML]{B2ED75}\textbf{94.79} & 90.10 & 57.81 & 5.73 & 1.56 & 52.34 & \cellcolor[HTML]{DBFFD2}94.53 \\
\textbf{VT} & 21.04 & 1.04 & \cellcolor[HTML]{B2ED75}\textbf{26.87} & 16.04 & 0.21 & 0.00 & 3.96 & \cellcolor[HTML]{DBFFD2}20.00 & 51.04 & 50.00 & \cellcolor[HTML]{B2ED75}\textbf{61.25} & 41.67 & 0.62 & 23.33 & 52.29 & \cellcolor[HTML]{DBFFD2}50.21 \\
\textbf{FWE} & 48.61 & 19.44 & 36.36 & 51.39 & \cellcolor[HTML]{B2ED75}\textbf{54.86} & 20.49 & 20.49 & \cellcolor[HTML]{DBFFD2}43.40 & 75.35 & 57.99 & 63.19 & 57.99 & \cellcolor[HTML]{DBFFD2}67.7 & 48.61 & 38.89 & \cellcolor[HTML]{B2ED75}\textbf{71.18} \\
\textbf{QA1} & 30.21 & 14.58 & 23.96 & \cellcolor[HTML]{B2ED75}\textbf{30.21} & 22.92 & 15.62 & 22.92 & \cellcolor[HTML]{DBFFD2}28.12 & 78.12 & \cellcolor[HTML]{DBFFD2}76.04 & 73.96 & 67.71 & 51.04 & 48.96 & \cellcolor[HTML]{B2ED75}\textbf{78.12} & \cellcolor[HTML]{DBFFD2}76.04 \\
\textbf{QA2} & 36.46 & 16.67 & 34.38 & \cellcolor[HTML]{DBFFD2}35.42 & 27.08 & 16.67 & 26.04 & \cellcolor[HTML]{B2ED75}\textbf{37.50} & 40.62 & 35.29 & 38.54 & 38.54 & 34.38 & 33.33 & \cellcolor[HTML]{DBFFD2}39.58 & \cellcolor[HTML]{B2ED75}\textbf{40.62} \\ \midrule
\textbf{AVG.} & 65.04 & 7.16 & 56.37 & \cellcolor[HTML]{DBFFD2}57.91 & 11.82 & 5.15 & 9.25 & \cellcolor[HTML]{B2ED75}\textbf{63.91} & 82.68 & \cellcolor[HTML]{DBFFD2}77.80 & 72.55 & 55.51 & 16.44 & 18.68 & 51.18 & \cellcolor[HTML]{B2ED75}\textbf{80.57} \\ \bottomrule
\end{tabular}

\caption{Accuracy results on RULER~\cite{bai2023longbench}. For Llama-2-7B-32K-Instruct, the context length is 32K and the token budget is 1024 (3.13\%). For Llama-3.1-8B-Instruct, the context length is 128K and the token budget is 2048 (1.56\%). For MagicPIG (MP), the token budget is approximately 2-3\% of the sequence length. \textbf{SL} denotes StreamingLLM, while \textbf{S-KV} refers to SnapKV.} 
\label{tab:ruler}
\end{table*}
}

In this section, we evaluate \algo's performance in terms of both accuracy and efficiency.

\subsection{Experimental Setup}
\noindent\textbf{Experiment platform.} We conduct experiments on a machine equipped with a 48GB HBM GPU delivering up to 149.7 TFLOPS (FP16) and 96 cores. The system runs Ubuntu~24.04, utilizing CUDA~12.1, PyTorch~2.4~\cite{pytorch}, FlashInfer~\cite{flashinfer}.

\noindent\textbf{Baselines and configurations.} We compare \algo with the state-of-the-art top-$k$ attention baselines: Loki~\cite{loki} (low-rank) and Quest~\cite{quest} (block-level), both of which are variants of top-$k$ attention. In addition, we further compare \algo with MagicPIG~\cite{MagicPig}, which accelerates top-$k$ attention through locality sensitive hashing (LSH)~\cite{LSH}. LSH is another kind of hashing method, which mainly utilizes random projections to generate hash codes. Different from learning-to-hash, LSH typically requires massive hash bits to ensure accuracy. More details about LSH can be seen in~\cite{LSH}. We also compare \algo with some KVCache compression methods, including StreamingLLM~\cite{streamllm}, H2O~\cite{h2o} and SnapKV~\cite{snapkv}.

We adopt the recommended configurations (e.g., channels, block size) from the original papers for all baselines. For \algo, we set rbit=128, a versatile configuration that maintains quality across most tasks. Following~\cite{quest}, we use vanilla attention for the first two layers, which are typically outlier layers in top-$k$ attention methods. 

We additionally add the vanilla transformer with full attention mechanism (denoted by dense) as a reference baseline to demonstrate the effectiveness and efficiency of \algo.

\noindent\textbf{Models and datasets.} We mainly evaluate \algo on two mainstream large language models: Llama2~\cite{llama2-7b-32k-instruct} and Llama3.1~\cite{llama3.1}. The test datasets include two widely used benchmarks: Longbench-e~\cite{bai2023longbench} and RULER~\cite{hsieh2024ruler}. LongBench-e is a multitask benchmark involving QA, document summarization, and code understanding. RULER focuses on retrieval tasks over extremely long contexts.

Due to space constraints, we only report selected results here. More results including more models and tasks are provided in  Appendix~\ref{appndx:exp}.

\subsection{Accuracy Evaluation}
\label{sec:evaluations:acc}

\noindent\textbf{Evaluation on LongBench-e.} 

First, we test all the methods on the LongBench-e tasks, which involve QA, document summarization, and code understanding. From Table~\ref{tab:longbench}, we observe that for both Llama2 and Llama3.1, \algo achieves results comparable to the vanilla full attention mechanism and outperforms all other baselines in most cases. 

\noindent\textbf{Evaluation on RULER.} Next, we test all the  methods on the long-context tasks. RULER can be used to construct retrieval, tracing, aggregation and QA tasks with any length. Note that the input sequence length should not surpass the maximum context window size of model. Hence, we  test Llama2 and Llama3.1 on 32k-long and 128k-long sequences, respectively.  We set the token budget as 1024 for Llama2 and 2048 for Llama3.1 (only 3.12\% and 1.56\% of total sequence length). The results shown in Table~\ref{tab:ruler} is in line with results test on Longbench-e. For long-context inference,  \algo can  still maintain the accuracy of the  vanilla  full attention mechanism, while all the other competitors has obvious accuracy degradation, which shows the superiority of \algo.

\subsection{Efficiency Evaluation}
\label{sec:evaluations:perf}

This subsection evaluates \algo's efficiency against baselines through three aspects: (1) end-to-end inference performance, (2) decoding efficiency analysis across different input scales, and (3) comparison with MagicPig using HATA's KVCache offloading extension. For Quest, we directly use their open-source high-performance implementation~\cite{quest-code}. For the full attention baseline (dense), we adopt the recently widely-used vLLM~\cite{vllm} implementation. For Loki, since it did not provide a high-performance implementation, here we give an efficient realization based on triton, which is detailed in Appendix~\ref{appndx:loki}. % Note that MagicPIG adopts the offloading technology to save the HMB memory, which will introduce additional latency, so directly comparing time efficiency is not fair. In Appendix~\ref{appndx:exp:offloading}, we give an offloading version of  \algo, named \algo-off, and compare \algo-off with MagicPIG.

\begin{figure}[t]
\begin{center}
\centerline{\includegraphics[width=\columnwidth]{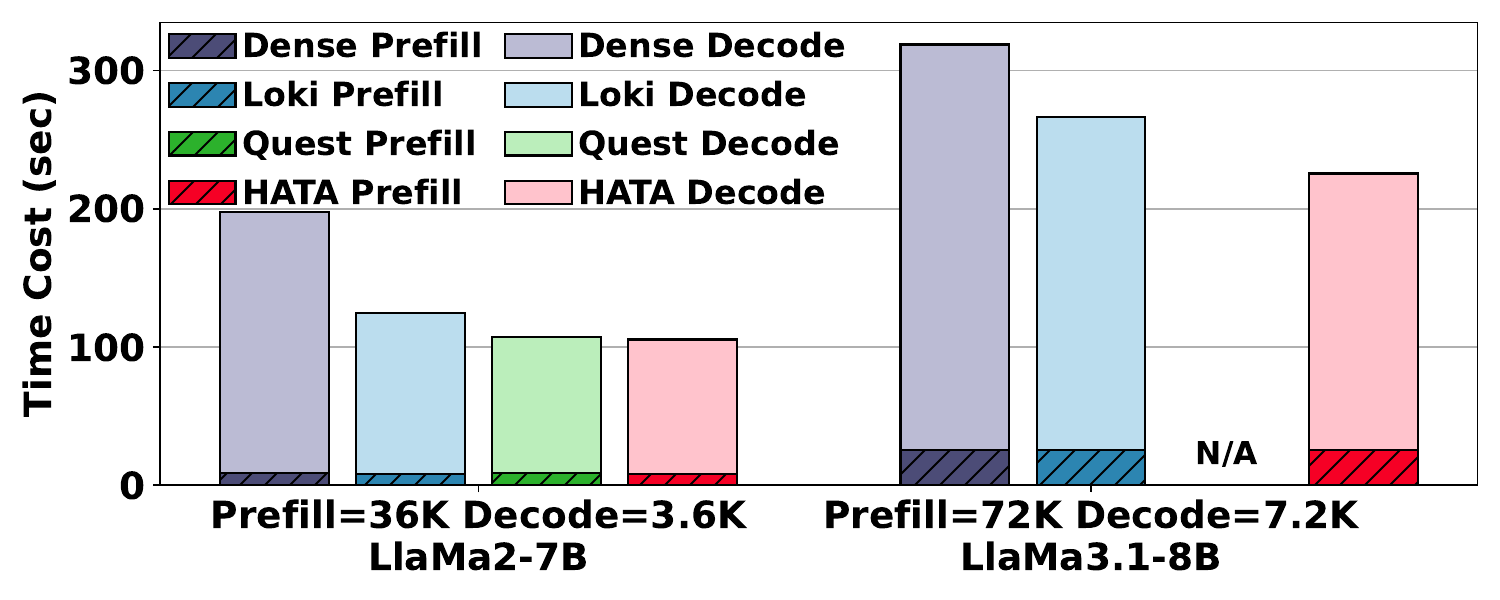}}
\caption{End-to-end performance comparison of LLM inference under 1.56\% token selection.}
\label{end-to-end-performance}
\end{center}
\end{figure}

\noindent\textbf{End-to-end inference efficiency.}
Both  \algo and the above-mentioned compared methods are designed for speeding up the LLM decoding. In Figure~\ref{end-to-end-performance}, we compare the decoding time cost of all the methods with the same sequence length. In addition, we also show the prefill time cost to measure the end-to-end efficiency performance of these methods comprehensively. Here we only report the time efficiency of Quest on Llama2, since its open-source high-performance implementation does not support GQA so far.

From Figure~\ref{end-to-end-performance}, we see that \algo, Loki, and Quest all have significant speedup in decoding compared with the vanilla attention mechanism, and among them, \algo achieves the highest decoding efficiency. On the other hand, we can see that for LoKi, Quest, and \algo, the prefill time is similar to the vanilla attention mechanism, so all of them can improve the end-to-end inference efficiency. Though it is expected that Quest can achieve similar time efficiency to \algo, \algo can achieve better accuracy under the same budget.

\begin{figure}[t]
\begin{center}
\centerline{\includegraphics[width=\columnwidth]{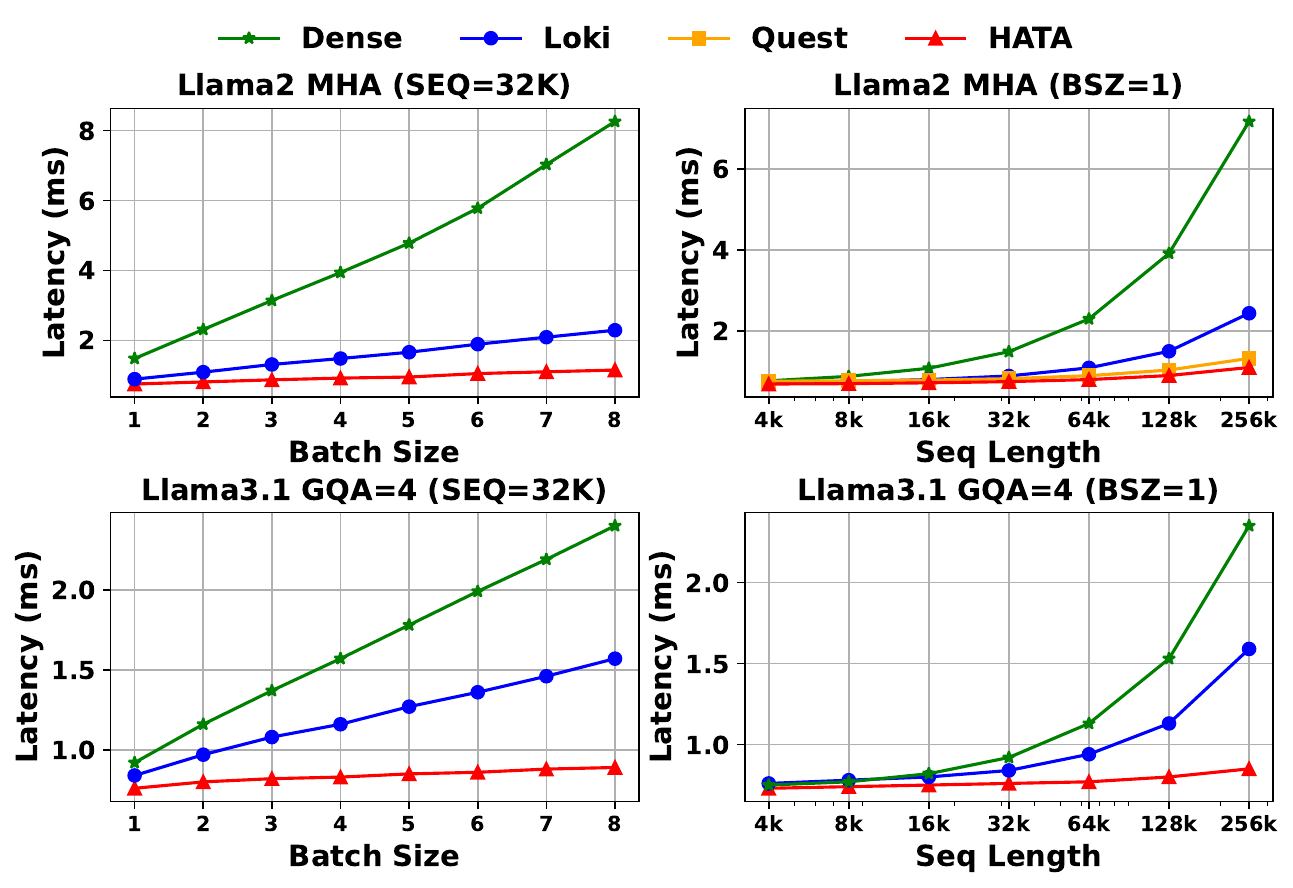}}
\caption{Performance comparison of a single transformer layer under 1.56\% token selection.}
\label{single-layer-performance}
\end{center}
\vspace{-5pt}
\end{figure}

{
\setlength{\tabcolsep}{2pt}
\begin{table}[h]
\centering
\small
\begin{tabular}{c|cc|cc}
\toprule
\multirow{2}{*}{\textbf{\begin{tabular}[c]{@{}c@{}}Time 
  Cost \\ (second) \end{tabular}}} & \multicolumn{2}{c|}{\textbf{Llama2}} & \multicolumn{2}{c}{\textbf{Llama3.1}} \\
                 & \textbf{MagicPig} & \textbf{\algo-off} & \textbf{MagicPig} & \textbf{\algo-off} \\ \midrule
\textbf{Prefill} & 49.89             & 8.26          & 33.24             & 25.09         \\
\textbf{Decode}  & 38.21             & 15.04         & 41.69             & 15.86         \\
\textbf{Total}   & 88.10             & 23.30         & 74.93             & 40.95         \\ \bottomrule
\end{tabular}
\caption{Offloading performance comparison between \algo-off and MagicPIG. We set the prefill length as 36K and 72K for \textbf{Llama2} and \textbf{Llama3.1} respectively, and the decode length is set as 500 for both model. For MagicPIG, the token budget is approximately 2-3\% of the sequence length, and for \algo-off we set the token budgets as 1.56\%.}
\label{tab:exp_offloading}
\end{table}
}

% \paragraph{Single layer decoding efficiency}
\noindent\textbf{Decoding efficiency across varying input scales.} We further evaluate \algo across varying batch sizes and input sequence lengths. Due to GPU memory constraints, we evaluate only a single transformer layer of Llama2 and Llama3.1. Since prefill costs are similar across baselines, we focus on decoding step latency. Furthermore, since the high-performance implementation of the open-source Quest is limited to a batch size of 1 and MHA models, we evaluate it solely across varying sequence lengths on Llama2. As shown in Figure~\ref{single-layer-performance}, \algo outperforms all the baselines. Notably, with longer sequences and larger batches, \algo achieves greater speedups. With batch size = 8 and sequence length = 32K, \algo reaches up to 7.20$\times$ speedup over Dense and 1.99$\times$ over Loki. At batch size = 1 and sequence length = 256K, \algo achieves up to 6.51$\times$ speedup over Dense, 2.21$\times$ over Loki and 1.19$\times$ over Quest. These results demonstrate \algo's high inference efficiency across tasks of varying scales.

\noindent\textbf{Efficiency with KVCache offloading}
For fair comparison with MagicPIG, we introduce HATA-off, an offloading variant of HATA inspired by InfiniGen~\cite{infinigen}. By combining KVCache offloading with prefetching, HATA-off reduces GPU memory usage while maintains inference efficiency. On Llama2 and Llama3.1 with PCIe 4.0 and 48 CPU threads, HATA-off achieves 6.04× (prefill) and 2.54× (decode) speedups over MagicPIG on Llama2, and 1.32× (prefill) and 2.63× (decode) on Llama3.1, as shown in Table~\ref{tab:exp_offloading}. The improvements come from: (1) eliminating MagicPIG's expensive LSH hashing (i.e., 1,500-bit hash bits for a 128D vector), and (2) our GPU-optimized attention with lightweight hashing and KV prefetching, surpassing MagicPIG's CPU-based method. These innovations enable scalable, memory-efficient long-sequence inference.
\section{Related Works}
\label{related_works}
Our work \algo advances top-$k$ attention for accelerating KVCache-enabled LLM inference, but significantly differs from existing top-$k$ attention methods. Prior top-$k$ attention methods~\cite{loki,sparq,infinigen,quest,infllm} assume precise qk score estimation is essential to replicate full attention, incurring high computational or memory overhead to minimize errors. Other hashing-based methods for LLMs fail to achieve practical inference acceleration. MagicPIG~\cite{MagicPig} employs locality-sensitive hashing but relies on high-bit representations, limiting speed and sacrificing accuracy. HashAttention~\cite{HashAttention}, a concurrent work, also uses learning-to-hash but adopts a custom training approach, lacks extensive testing across datasets and models, and overlooks system challenges in applying hashing to top-$k$ attention. Some works~\cite{sun2021sparse-learning-to-hash} attempt hashing in LLM training but fail to transfer it to inference due to fundamental differences between the two phases.

Other orthogonal approaches focus on compressing KVCache content. Eviction methods~\cite{h2o,keyformer} remove less important tokens but risk information loss and dynamic token importance shifts, potentially degrading output quality. Quantization methods~\cite{kivi,kvquant} compress the KVCache, though their speedup gains are limited by low compression ratios.

Finally, the offloading methods~\cite{infinigen, flexgen, shadowkv} transfer KVCache to CPU memory to reduce HBM memory usage. \algo is orthogonal to these methods and can be combined with them. We developed \algo-off to demonstrate how \algo can be efficiently combined with KVCache offloading without compromising performance.
\section{Conclusion}
We introduced Hash-Aware Top-$k$ Attention (\algo), a hardware-efficient method for faster LLM inference. \algo offers a systematic exploration and validation of the integration of learning-to-hash techniques into top-$k$ attention mechanisms, achieving up to 7.2$\times$ speedup over dense attention and outperforming SOTA methods in accuracy and performance, establishing it as an effective solution for LLM inference acceleration. 
\section{Limitations}
With learning-to-hash, \algo has achieved notable success in top-$k$ attention. However, it still has the following limitations:

\noindent\textbf{Larger-scale training.}  
\algo’s training data consists of millions of query-key pairs sampled from a limited number of sequences, which is sufficient to train effective hash weights. However, expanding the diversity and scale of the training data could further enhance the quality of the hash weights. We plan to explore this in the future to improve \algo’s performance across a wider range of tasks.

\noindent\textbf{Fields of application.}  
\algo is designed to accelerate LLM inference with long contexts or large batch sizes. For small batch sizes and short context sequences, \algo does not provide significant speedup, as the attention module is not the bottleneck in these cases. 

\noindent\textbf{MLA adaptor.}  
Over the past month, Multi-Latent Head Attention (MLA) in DeepSeek~\cite{liu2024deepseek} has gained significant attention. While we’ve evaluated \algo on MHA and GQA tasks, it remains untested with MLA, which we leave as future work.

\section*{Acknowledgements}
 We thank the anonymous reviewers for their insightful comments. This work is supported by the Strategic Priority Research Program of the Chinese Academy of Sciences, Grant No. XDB0660101, XDB0660000, XDB0660100, and Huawei Technologies.

\bibliography{ref}

\clearpage
\appendix

\appendix

\begin{table*}[t]
\centering
\small
\begin{tabular}{@{}llll@{}}
\toprule
\multicolumn{1}{c|}{\textbf{Model}} &  \multicolumn{1}{c|}{\textbf{Abbr.}} &\multicolumn{1}{c|}{\textbf{Configs}} & \multicolumn{1}{c}{\textbf{Values}} \\ \midrule
\multicolumn{1}{l|}{\multirow{5}{*}{Llama-2-7B-32K-Instruct~\cite{llama2-7b-32k-instruct}}}  &  \multicolumn{1}{l|}{\multirow{5}{*}{Llama2}} & \multicolumn{1}{l|}{\#Layer}             & 32      \\
\multicolumn{1}{l|}{}  &  \multicolumn{1}{l|}{}  &\multicolumn{1}{l|}{\#Attention Heads}   & 32      \\
\multicolumn{1}{l|}{}  &  \multicolumn{1}{l|}{}  &\multicolumn{1}{l|}{\#KV Heads}          & 32      \\
\multicolumn{1}{l|}{}  &  \multicolumn{1}{l|}{}  &\multicolumn{1}{l|}{Hidden Size}         & 4096    \\
\multicolumn{1}{l|}{}  &  \multicolumn{1}{l|}{}  &\multicolumn{1}{l|}{Max Context Length} & 32768   \\ \midrule
\multicolumn{1}{l|}{\multirow{5}{*}{Llama-3.1-8B-Instruct~\cite{llama3.1}}}    &  \multicolumn{1}{l|}{\multirow{5}{*}{Llama3.1}} &\multicolumn{1}{l|}{\#Layer}             & 32      \\
\multicolumn{1}{l|}{}  &  \multicolumn{1}{l|}{}  &\multicolumn{1}{l|}{\#Attention Heads}   & 32      \\
\multicolumn{1}{l|}{}  &  \multicolumn{1}{l|}{}  &\multicolumn{1}{l|}{\#KV Heads}          & 8       \\
\multicolumn{1}{l|}{}  &  \multicolumn{1}{l|}{}  &\multicolumn{1}{l|}{Hidden Size}         & 4096    \\
\multicolumn{1}{l|}{}  &  \multicolumn{1}{l|}{}  &\multicolumn{1}{l|}{Max Context Length} & 131072  \\ \midrule
\multicolumn{1}{l|}{\multirow{5}{*}{Qwen2.5-14B-Instruct-1M~\cite{qwen2.5-1m}}} &  \multicolumn{1}{l|}{\multirow{5}{*}{Qwen2.5-14B}} &\multicolumn{1}{l|}{\#Layer}             & 48      \\
\multicolumn{1}{l|}{}  &  \multicolumn{1}{l|}{}  &\multicolumn{1}{l|}{\#Attention Heads}   & 40      \\
\multicolumn{1}{l|}{}  &  \multicolumn{1}{l|}{}  &\multicolumn{1}{l|}{\#KV Heads}          & 8       \\
\multicolumn{1}{l|}{}  &  \multicolumn{1}{l|}{}  &\multicolumn{1}{l|}{Hidden Size}         & 5120    \\
\multicolumn{1}{l|}{}  &  \multicolumn{1}{l|}{}  &\multicolumn{1}{l|}{Max Context Length} & 1010000 \\ \midrule
\multicolumn{1}{l|}{\multirow{5}{*}{Qwen2.5-32B-Instruct~\cite{qwen2.5}}}    &  \multicolumn{1}{l|}{\multirow{5}{*}{Qwen2.5-32B}} &\multicolumn{1}{l|}{\#Layer}             & 64      \\
\multicolumn{1}{l|}{}  &  \multicolumn{1}{l|}{}  &\multicolumn{1}{l|}{\#Attention Heads}   & 40      \\
\multicolumn{1}{l|}{}  &  \multicolumn{1}{l|}{}  &\multicolumn{1}{l|}{\#KV Heads}          & 8       \\
\multicolumn{1}{l|}{}  &  \multicolumn{1}{l|}{}  &\multicolumn{1}{l|}{Hidden Size}         & 5120    \\
\multicolumn{1}{l|}{}  &  \multicolumn{1}{l|}{}  &\multicolumn{1}{l|}{Max Context Length} & 131072 \\ \bottomrule
\end{tabular}
\caption{Configurations of the models we used for evaluation.}
\label{tab:model-details}
\end{table*}

\begin{table*}[t]
\centering
\small
\begin{tabular}{l|l|l}
\toprule
\textbf{Methods} &  Abbr.&\textbf{Settings} \\
\midrule
Dense &  Dense&Inference with the full KVCache (dense attention) \\
\topk &  topk&Exact top-$k$ attention \\
Loki~\cite{loki} &  Loki&Top-$k$ attention, Number of channels = 32 \\
Quest~\cite{quest} &  Quest&Top-$k$ attention, Block size = 32 \\
MagicPIG~\cite{MagicPig} &  MP&Top-$k$ attention with KVCache Offloading, K=10, L=150 \\
StreamingLLM~\cite{streamllm} &  SL&KVCache compression, number of attention sinks=4 \\
H2O~\cite{h2o} &  H2O&KVCache compression, heavy hitter ratio=recent ratio \\
SnapKV~\cite{snapkv} &  S-KV&KVCache compression, length of observation window=16 \\
\algo &  HATA&Top-$k$ attention, trained hash weights (128 bits) \\
\algo-off &  HATA-off& \algo with KVCache offloading, trained hash weights (128 bits) \\
\bottomrule
\end{tabular}
\caption{Configurations for the evaluated methods.}
\label{tab:appdnx_methods_setting}
\end{table*}

{\setlength{\tabcolsep}{2.3pt}
\begin{table*}[ht]
\centering
\begin{tabular}{
l|ccccccccc|c}
\toprule
\textbf{Methods} & \textbf{Sum} & \textbf{Choice} & \textbf{BookQA} & \textbf{DialQA} & \textbf{ZhQA} & \textbf{NumStr} & \textbf{Passkey} & \textbf{Debug} & \textbf{MathFind} & \textbf{AVG.}\\
\midrule
\textbf{Dense} & 20.36 & 57.64 & 38.33 & 18.50 & 27.57 & 97.80 & 100.00 & 22.59 & 23.71 & 45.17\\
\midrule
\textbf{\algo} & 19.27 & 57.64 & 37.52 & 18.50 & 27.27 & 96.44 & 100.00 & 22.59 & 23.71 & 44.77\\
\bottomrule
\end{tabular}
\caption{Accuracy results on \textbf{InfiniteBench}~\cite{zhang2024infinitebench} for \textbf{Llama3.1} model with sparse token budget=2048. Samples exceeding the model's maximum context window are truncated to fit within it.}
\label{tab:appndx_infinitebench_llama31}

\vspace{0.1cm}

\centering
\begin{tabular}{
l|cccccc|c}
\toprule
\textbf{Methods} & \textbf{Easy.Short} & \textbf{Easy.Medium} & \textbf{Easy.Long} & \textbf{Hard.Short} & \textbf{Hard.Medium} & \textbf{Hard.Long} & \textbf{Total}\\
\midrule
\textbf{Dense} & 44.07 & 28.41 & 31.11 & 32.23 & 25.98 & 25.40 & 30.42 \\
\textbf{\topk} & 40.68 & 25.00 & 33.33 & 29.75 & 25.20 & 23.81 & 28.63 \\
\midrule
\textbf{\algo} & 38.98 & 27.27 & 35.56 & 29.75 & 26.77 & 25.40 & 29.62 \\
\bottomrule
\end{tabular}
\caption{Accuracy results on \textbf{LongBench-v2}~\cite{bai2024longbenchv2} for \textbf{Llama3.1} model with sparse token budget=1024. Samples exceeding the model's maximum context window are truncated to fit within it.}
\label{tab:appndx_longbenchv2_llama31}
\end{table*}
}

\section{Additional Evaluation Results}
\label{appndx:exp}

In this section, we present supplemental evaluation results.

\begin{itemize}
    \item In~\ref{appndx:exp:models and baselines}, we provide detailed configurations of the models and top-k attention algorithm baselines used for evaluation.
    \item In~\ref{appndx:exp:additional accuracy results}, we additionally compare \algo with dense model in three commonly used benchmarks (InfiniBench, NIAH and LongBench-v2).
    \item In~\ref{appndx:exp:ablation_study}, we conduct ablation studies on \algo, analyzing the effects of hash bits and token budget on inference accuracy, as well as the efficiency gains achieved through the optimizations discussed in Sec~\ref{sec:kernel-impl}.
    \item In~\ref{appndx:exp:larger_tasks}, we show that \algo can successfully scale to larger models (Qwen2.5-14B and Qwen2.5-32B) and handle longer contexts (up to 256K tokens).
\end{itemize}

\subsection{Models and Baselines}
\label{appndx:exp:models and baselines}

Table~\ref{tab:model-details} summarizes key parameters of the evaluated models. Llama2 uses multi-head attention (MHA), while the other three employ group-query attention (GQA). Table~\ref{tab:appdnx_methods_setting} lists configurations of all attention methods used for comparison.

\subsection{Addtional Accuracy Results}
\label{appndx:exp:additional accuracy results}

We additionally test \algo across three commonly used benchmarks: InfiniBench~\cite{zhang2024infinitebench}, LongBench-v2~\cite{bai2024longbenchv2} and Need-in-a-Haystack~\cite{Needle-in-a-Haystack}. In all the three benchmarks, \algo achieves near-lossless accuracy compared with dense model.

\noindent\textbf{InfiniteBench.} InfiniteBench covers tasks of QA, coding, dialogue, summarization, and retrieval, with an average length of 214K. We evaluated \algo on this benchmark using Llama3.1 to demonstrate its effectiveness in complex long-context scenarios. The results are shown in Table~\ref{tab:appndx_infinitebench_llama31}.

\noindent\textbf{LongBench-v2.}
LongBench-v2 is an update of the LongBench benchmark, which comprises 503 multiple-choice questions with context lengths spanning from 8K to an extensive 2M words. We employed the Llama3.1 model on LongBench-v2. The accuracy results are categorized based on two key dimensions: task difficulty (Easy, Hard) and context length (Short, Medium, Long). As shown in Tabel~\ref{tab:appndx_longbenchv2_llama31}, \algo consistently maintains model accuracy across most tasks, and even outperforms the exact top-$k$ attention in certain scenarios.

\begin{figure}[t]
\begin{center}
\centerline{\includegraphics[width=.995\columnwidth]{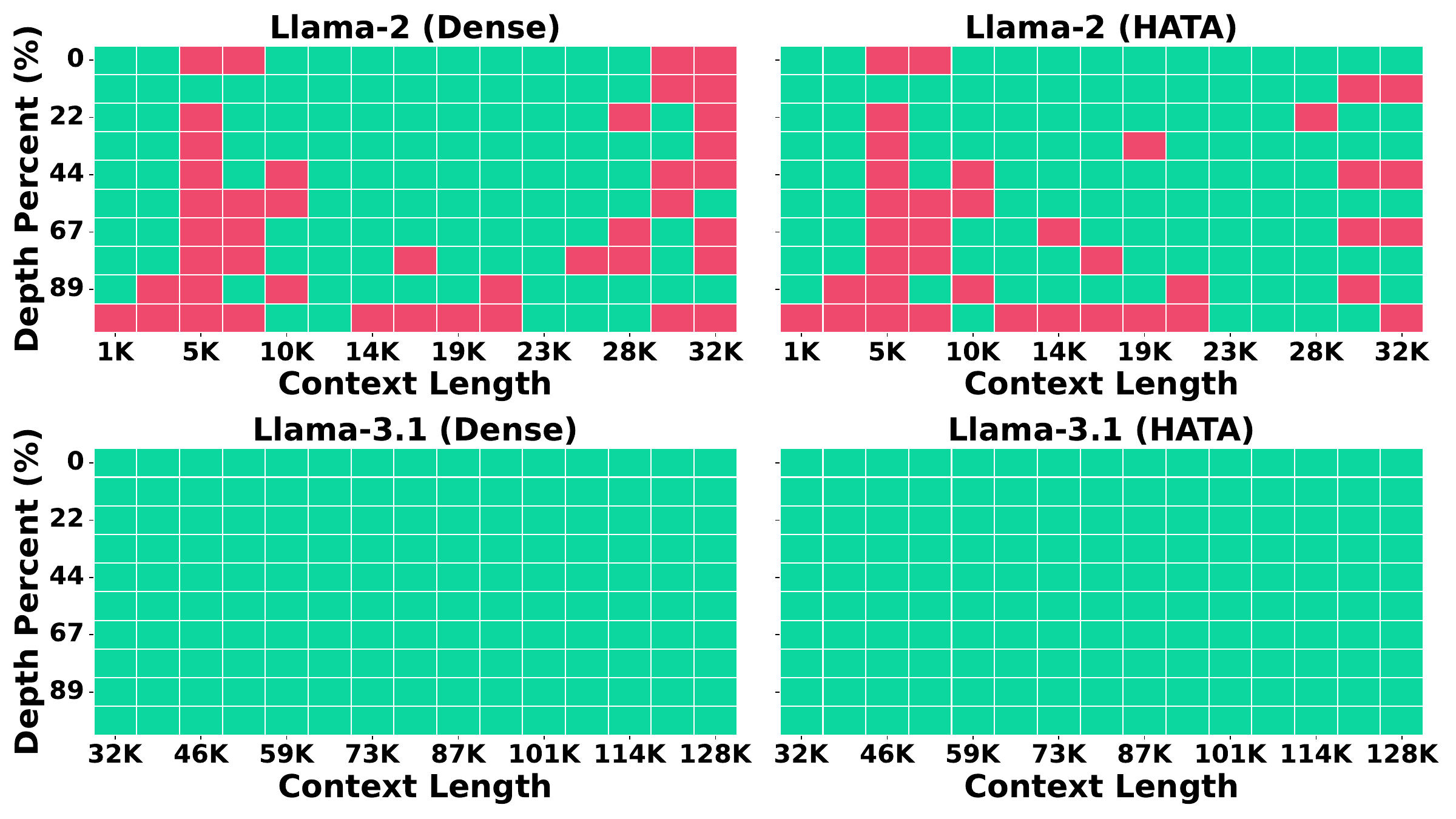}}
\caption{Needle-in-a-Haystack evaluation results. For \algo, the sparse token budget is 512 for Llama2 and 2048 for Llama3.1.}
\label{fig:niah}
\end{center}
\end{figure}

\noindent\textbf{Needle-in-a-Haystack.} Needle-in-a-Haystack is a retrieval task. By varying the haystack length and the depth of the needle, we can comprehensively evaluate the effectiveness of \algo in retrieval tasks. For Llama2, we set the haystack length ranging from 1K to 32K to fit within the model's context window. While for Llama3.1, we extended the range from 32K to 128K. As shown in Figure~\ref{fig:niah}, \algo achieves accuracy results similar to the dense attention.

\subsection{Ablation Study}
\label{appndx:exp:ablation_study}

In this subsection, we conduct ablation studies on \algo. For accuracy, we investigate the impact of sparse token budget and the number of hash bits on \algo's performance. For inference efficiency, we examine the performance improvements brought by the optimization introduced in Sec~\ref{sec:kernel-impl}.

\begin{figure}[t]
\begin{center}
\centerline{\includegraphics[width=.995\columnwidth]{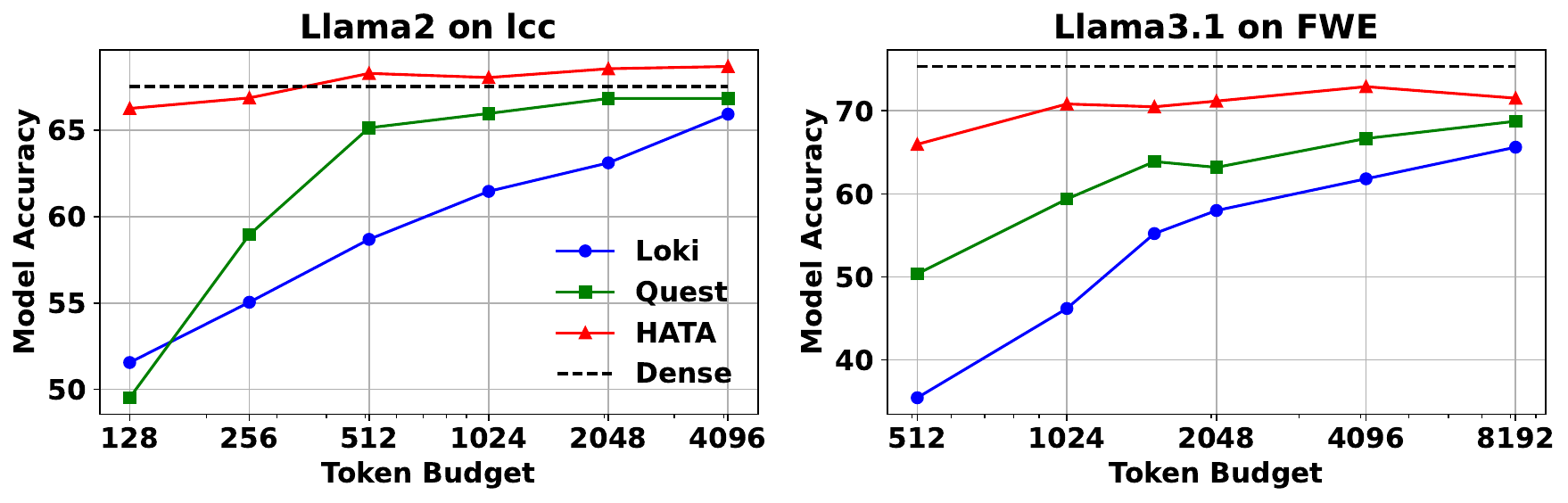}}
\caption{Token budget ablation.}
\label{cache-budget}
\end{center}
\end{figure}

\noindent\textbf{Token budget ablation.} First, we examine the impact of token budgets on \algo's performance. As shown in Figure~\ref{cache-budget}, \algo consistently outperforms Quest and Loki under the various budgets. Notably, as budgets decrease, \algo's accuracy degrades minimally, maintaining acceptable performance even under 0.4\% token ratio, highlighting the strong potential of learning-to-hash.

\begin{figure}[t]
\begin{center}
\centerline{\includegraphics[width=.995\columnwidth]{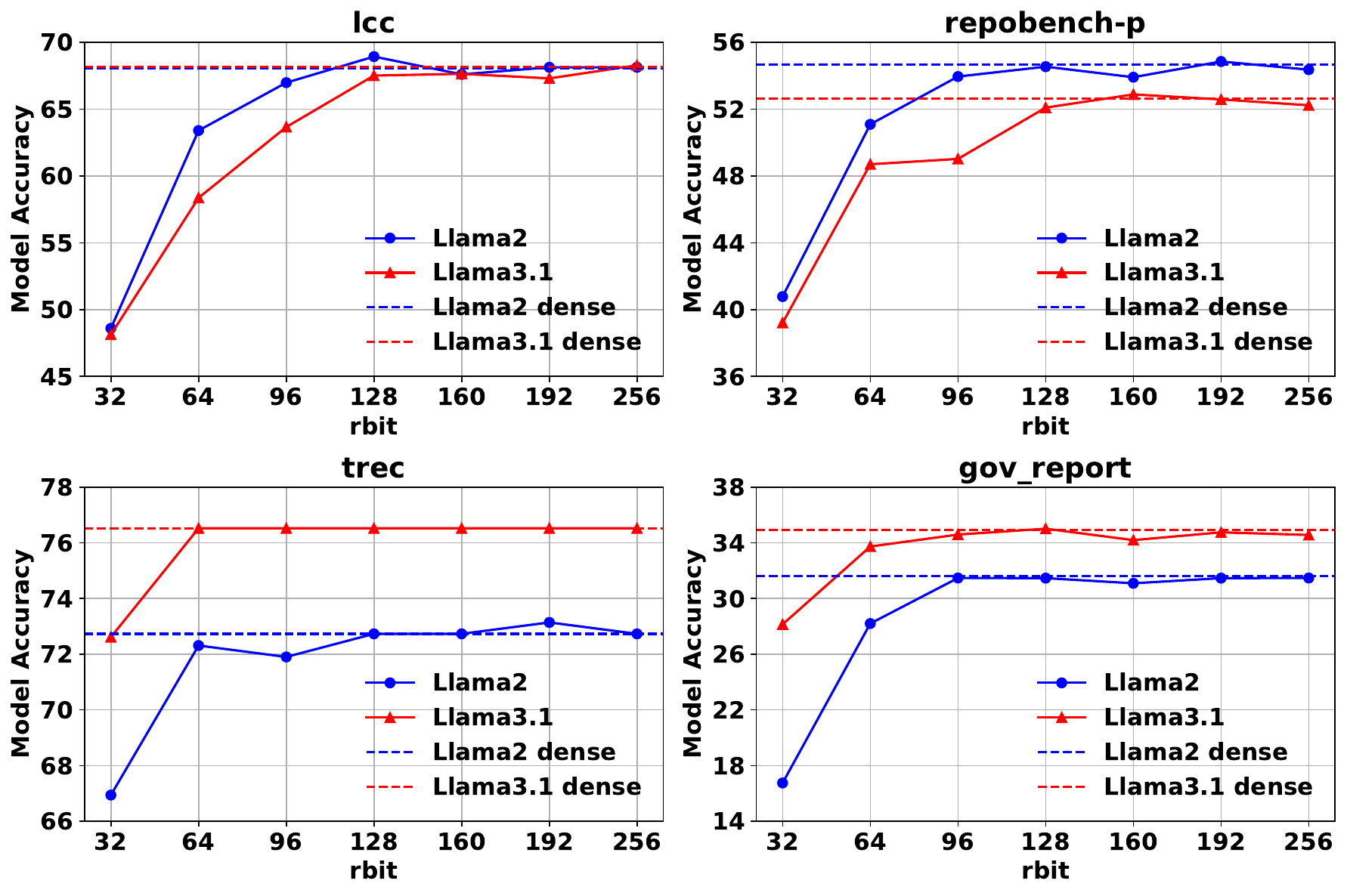}}
\caption{Hash bits ablation.}
\label{fig:appndx_rbits_eblation}
\end{center}
\end{figure}

\noindent\textbf{Hash bits ablation.} Next, we explore the effect of hash bit count (rbit) on inference accuracy. As depicted in Figure~\ref{fig:appndx_rbits_eblation}, increasing rbit from 32 to 128 leads to improved accuracy across four datasets and two models. At rbit=128, accuracy approaches near-lossless levels, comparable to dense attention, with further increases causing only minor fluctuations. Therefore, we adopt rbit=128 as an optimal setting, balancing accuracy and computational efficiency.

\begin{figure}[h]
\begin{center}
\centerline{\includegraphics[width=.995\columnwidth]{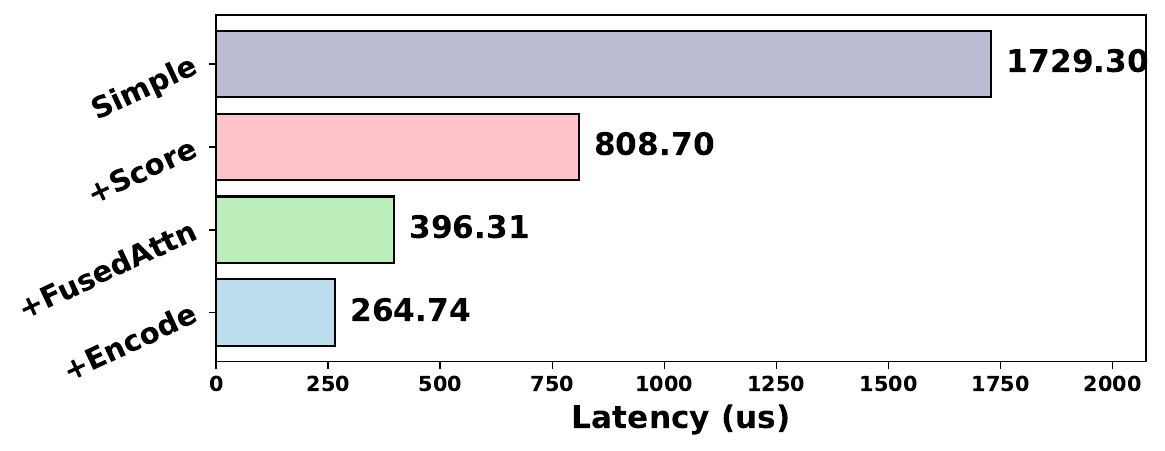}}
\caption{Performance ablation study of \algo optimizations under 1.56\% token budget.}
\label{performance-ablation}
\end{center}
\end{figure}

\noindent\textbf{Optimizations ablation.} Lastly, we evaluate the impact of \algo's optimizations on inference efficiency: high-performance hamming score operator (\textbf{Score}), fused gather with FlashAttention (\textbf{FusedAttn}), and kernel fusion for hash encoding (\textbf{Encode}). Using Llama2's attention module with 128K input, we apply these optimizations incrementally. Figure~\ref{performance-ablation} shows that \textbf{Score} reduces the total latency of attention module by 53.2\%, \textbf{FusedAttn} by 23.8\%, and \textbf{Encode} by 7.6\%. The fully-optimized \algo achieves a 6.53$\times$ speedup over a simple PyTorch implementation.

\begin{table*}[ht]
{\setlength{\tabcolsep}{1.7pt}

\centering
\begin{tabular}{l|ccccccccccccc|c}
\toprule
\textbf{Methods} & \textbf{LCC} & \textbf{PRetr} & \textbf{HQA} & \textbf{TQA} & \textbf{Repo} & \textbf{Sam} & \textbf{Trec} & \textbf{MQA} & \textbf{2Wiki} & \textbf{Gov} & \textbf{PCnt} & \textbf{MltN} & \textbf{Qaspr} & \textbf{AVG.} \\
\midrule
\textbf{Dense} & 44.32& 100.00 & 65.96 & 88.41 & 36.25 & 45.52 & 76.34 & 53.73 & 60.68 & 31.93 & 22.83 & 22.14 & 41.41 & 53.04 \\
\midrule
\textbf{\algo} & 44.86 & 99.67 & 65.87 & 88.49 & 37.41 & 45.41 & 76.67 & 53.45 & 60.70 & 31.25 & 20.50 & 22.02 & 41.46 & 52.90 \\
\bottomrule
\end{tabular}
\caption{Accuracy results on \textbf{LongBench-e}~\cite{bai2023longbench} for \textbf{Qwen2.5-14B-Instruct-1M}~\cite{qwen2.5-1m} model with sparse token budget=512.}
\label{tab:appndx_longbench_qwen14b1m}
% \end{table*}
}

{\setlength{\tabcolsep}{1.7pt}
% \begin{table*}[hb]
\centering
\begin{tabular}{l|ccccccccccccc|c}
\toprule
\textbf{Methods} & \textbf{LCC} & \textbf{PRetr} & \textbf{HQA} & \textbf{TQA} & \textbf{Repo} & \textbf{Sam} & \textbf{Trec} & \textbf{MQA} & \textbf{2Wiki} & \textbf{Gov} & \textbf{PCnt} & \textbf{MltN} & \textbf{Qaspr} & \textbf{AVG.} \\
\midrule
\textbf{Dense} & 54.04& 99.83 & 69.27 & 86.26 & 36.03 & 43.60 & 75.67 & 52.28 & 60.69 & 30.14 & 22.00 & 21.91 & 44.08 & 53.52 \\
\midrule
\textbf{\algo} & 53.90 & 100.00 & 68.58 & 87.55 & 36.22 & 42.75 & 75.67 & 52.29 & 60.51 & 30.17 & 22.00 & 21.79 & 43.70 & 53.47 \\
\bottomrule
\end{tabular}
\caption{Accuracy results on \textbf{LongBench-e}~\cite{bai2023longbench} for \textbf{Qwen2.5-32B-Instruct}~\cite{qwen2.5} model with sparse token budget=512.}
\label{tab:appndx_longbench_qwen32b}
% \end{table*}
}

{\setlength{\tabcolsep}{2.9pt}
% \begin{table*}[hb]
\centering
\begin{tabular}{l|ccccccccccc|c}
\toprule
\textbf{Methods} & \textbf{NS1} & \textbf{NS2} & \textbf{NS3} & \textbf{NMK1} & \textbf{NMK2} & \textbf{NMV} & \textbf{NMQ} & \textbf{VT} & \textbf{FWE} & \textbf{QA1} & \textbf{QA2} & \textbf{AVG.} \\
\midrule
\textbf{Dense} & 100.00 & 100.00 & 100.00 & 100.00 & 90.00& 85.00& 97.50& 100.00& 95.00& 60.00& 40.00& 87.95\\
\textbf{\topk} & 100.00& 100.00& 100.00& 100.00 & 90.00& 81.25 & 98.75 & 100.00 & 88.33 & 60.00& 40.00&87.12\\
\midrule
\textbf{\algo} & 100.00 & 100.00 & 100.00& 100.00 & 95.00& 85.00& 97.50& 96.00& 85.00& 60.00& 45.00& 88.05\\
\bottomrule
\end{tabular}
\caption{Accuracy results on \textbf{RULER(256K)}~\cite{bai2023longbench} for \textbf{Qwen2.5-14B-Instruct-1M}~\cite{qwen2.5-1m} model with sparse token budget=4096 (1.56\%)}
\label{tab:appndx_ruler256k_qwen14b1m}

}
\end{table*}

\subsection{Scalability to Larger-Scale Tasks}
\label{appndx:exp:larger_tasks}

We further scale \algo to larger models (14B and 32B) and longer context inputs (256K).

We assessed \algo's accuracy on \textbf{Qwen2.5-14B} and \textbf{Qwen2.5-32B} using LongBench-e, as detailed in Table~\ref{tab:appndx_longbench_qwen14b1m} and Table~\ref{tab:appndx_longbench_qwen32b}, respectively. For both 14B and 32B models, \algo maintains near-lossless accuracy, underscoring its efficacy with large-scale models.

We further evaluated \algo's performance on extreme-long contexts using RULER-256K on Qwen2.5-14B-Instruct-1M. The results, as shown in the table, demonstrate that \algo achieves comparable accuracy to exact top-$k$ attention, highlighting its capability to handle ultra-long context inputs effectively.

\newpage
\section{Configuration Details of \algo Training}
\label{appndx:hashtrain}

\subsection{Data Sampling}
\label{appndx:hashtrain:data}
We trained hash weights based on query and key data sampled from real world datasets. Detailed sampling steps for a given sequence are as follows:

\begin{enumerate}
    \item For a given token sequence of length $n$, generate its query $Q:=[q_1,q_2\dots q_n] \in \mathbb{R}^{n\times d}$ and key $K:=[k_1,k_2\dots k_n] \in \mathbb{R}^{n\times d}$ by prefilling.
    \item Randomly sample one query $q_m \in \mathbb{R}^{1\times d}, m\in[\lfloor \frac{n}{2} \rfloor,n)$, and then accordingly sample all the keys that comply with the causal constraint: $K_{m}=[k_1,k_2\dots k_m] \in \mathbb{R}^{m\times d}$. Then we form $m$ qk pairs $\{(q_m,k_1),(q_m,k_2)\dots (q_m,k_m)\}$.
    \item Compute qk score $Score=q_mK_m^T \in \mathbb{R}^{1\times m}$ and sort it in descending order.
    \item Split the qk pairs into positive and negative samples and assign similarity labels:

    For the qk pairs whose score lies in top 10\% of sorted $Score$, we view them as positive samples. They are assigned linearly decayed labels in $[1, 20]$;

    For the qk pairs whose score lies in bottom 90\% of sorted $Score$, we view them as negative samples, and assign fixed $-1$ as their similarity labels.

    \item Finally, we get $m$ triplets:
    
    $\{(q_m, k_1, s_1),(q_m, k_2, s_2),\dots,(q_m, k_m, s_m)\}$
    
    where $s_i$ is the similarity label we assigned in the previous step. A triplet is a basic unit for training. These triplets are independent of each other during training. They can be arbitrarily combined or shuffled along with data sampled from other sequences, which will help improve the generalization of training and avoid overfitting.
\end{enumerate}

After introducing how to collect samples from a single sequence, we clarify from where the sequences are sampled:

\begin{itemize}
    \item 5 samples from Qasper of LongBench~\cite{bai2023longbench} for short sequences;
    \item 2 samples each from LSHT and RepoBench-P of LongBench for medium-length sequences;
    \item 2 samples from LongBench-v2~\cite{bai2024longbenchv2} for ultra-long sequences.
\end{itemize}

The sampled sequences cover diverse domains including Chinese and English QA, code understanding, ensuring the diversity of training data.

To fit within the model's context window, we truncated some long sequences. The final training set for each model comprises 150K–300K qk pairs.

\subsection{Training Setup}
\label{appndx:hashtrain:settings}

In this section, we report the detailed settings of hash training. Firstly, in Table~\ref{tab:training_hyperparams}, we detail the hyperparameter values during training, which are shared by all the models.

During training, in order to facilitate data IO and shuffling, we organize the sampled data into chunks of 32K size. In each epoch, several chunks (2 for Llama2 and 3 for Llama3.1, Qwen2.5-14B, Qwen2.5-32B) will be loaded for training. Each training epoch will perform multiple iterations on these data. For all the models, 15 epochs and 20 iterations are required to train one layer's hash weights.

\begin{table}[ht]
\centering
\begin{tabular}{c|c|c}
\toprule
\multicolumn{1}{c|}{\textbf{Class}} & \multicolumn{1}{c|}{\textbf{\begin{tabular}[c]{@{}c@{}}Hyper-\\parameter\end{tabular}}} &  \multicolumn{1}{c}{\textbf{Value}} \\
\toprule
\multirow{4}{*}{\begin{tabular}[l]{@{}c@{}}Custom\\Hyperparamters\end{tabular}}
 & $\bm{\sigma}$ & $0.1$ \\
 & $\bm{\epsilon}$ & $0.01$ \\
 & $\bm{\lambda}$ & $1.0$ \\
 & $\bm{\eta}$ & $2.0$ \\
\toprule
\multicolumn{1}{l|}{\multirow{3}{*}{\begin{tabular}[l]{@{}l@{}}SGD Optimizer\\Hyperparamters\end{tabular}}} & LR & $0.1$ \\
\multicolumn{1}{l|}{} & Weight decay & $10^{-6}$ \\
\multicolumn{1}{l|}{} & Momentum & $0.9$ \\
\bottomrule
\end{tabular}
\caption{Hyperparameter values during hash training.}
\label{tab:training_hyperparams}
\end{table}

\section{High-Performance Implementation for Loki}
\label{appndx:loki}

As explained in Sec~\ref{sec:evaluations:perf}, Loki~\cite{loki} lacks a high-performance implementation. While Loki has provided a kernel fusion of gather and matrix multiplication, their implementation neither integrates with the widely-used FlashAttention2 kernels~\cite{flash-attention-2} nor provides efficient end-to-end inference, preventing fair performance comparisons. To address these limitations, we develop a high-performance Loki implementation with these optimizations:

\noindent\textbf{Fuse gather with FlashAttention.} We employ the identical high-performance fused gather-FlashAttention kernel for Loki as described in Sec~\ref{sec:kernel-impl}, ensuring fair comparison.

\noindent\textbf{High-performance score operator.} Similar to \algo's high-performance hamming score operator (see Sec~\ref{sec:kernel-impl}), we implemented an optimized scoring operator for Loki. This triton-based~\cite{triton} kernel computes approximate scores for token selection using the first $R$ channels of PCA-projected query and key vectors, eliminating the redundant memory access overhead of low-rank queries and keys.

\noindent\textbf{Static KVCache.} Static KVCache refers to a pre-allocated GPU memory space for storing key-value pairs. During a decoding step, this approach only requires copying the newly generated key-value pair into the allocated space, eliminating the costly tensor concatenation operation, which involves heavy data copy.

\end{document}